\documentclass[runningheads]{llncs}
\usepackage{subfig}
\usepackage{xcolor}
\usepackage{amsmath, amssymb}
\usepackage{graphicx}%
\usepackage{booktabs}
\usepackage[table,xcdraw]{xcolor}
\usepackage[T1]{fontenc}
\usepackage{graphicx,verbatim}
\begin{document}
%

\title{Towards Patient-Specific Deformable Registration in Laparoscopic Surgery}
%
\author{Alberto Neri\inst{1,2,3} \and
Veronica Penza\inst{1} \and
Nazim Haouchine\inst{3} \and Leonardo S. Mattos\inst{1}
\authorrunning{A. Neri et al.}
%
\institute{Biomedical Robotics Lab, Istituto Italiano di Tecnologia, Genoa, Italy \and
Department of Computer Science, Bioengineering, Robotics and Systems Engineering (DIBRIS), University of Genoa, Italy \and
Harvard Medical School, Brigham and Women’s Hospital, Boston, USA\\
\email{alberto.neri@iit.it}}
}

\maketitle              
\begin{abstract}
Unsafe surgical care is a critical health concern, often linked to limitations in surgeon experience, skills, and situational awareness. Integrating patient-specific 3D models into the surgical field can enhance visualization, provide real-time anatomical guidance, and reduce intraoperative complications. However, reliably registering these models in general surgery remains challenging due to mismatches between preoperative and intraoperative organ surfaces—such as deformations and noise.
To overcome these challenges, we introduce the first patient-specific non-rigid point cloud registration method, which leverages a novel data generation strategy to optimize outcomes for individual patients. Our approach combines a Transformer encoder-decoder architecture with overlap estimation and a dedicated matching module to predict dense correspondences, followed by a physics-based algorithm for registration. Experimental results on both synthetic and real data demonstrate that our patient-specific method significantly outperforms traditional agnostic approaches, achieving 45\% Matching Score with 92\% Inlier Ratio on synthetic data, highlighting its potential to improve surgical care.

\keywords{Point cloud registration  \and Image-guided Surgery.}

\end{abstract}

\section{Introduction}


Unsafe surgical care remains a significant global health concern, contributing to patient complications, prolonged hospital stays, increased healthcare costs, and potential disability or death \cite{safety2009guidelines}. Adverse events are often related to the experience, skills, and situational awareness of surgeons, particularly in recognizing anatomical variations and integrating preoperative data intraoperatively \cite{VISSER201558}. Enhancing surgical awareness by overlaying patient-specific 3D models - such as organs, tumors, and vessels -  onto the surgical field, can improve visualization and real-time anatomical guidance and reduce intraoperative complications \cite{BERNHARDT201766}. 
The application in general surgery is particularly challenging due to the dynamic nature of organ tissues. 
Factors like patient positioning, pneumoperitoneum insufflation, and physiological movements cause intraoperative organ surfaces to differ from preoperative imaging, making model-to-image registration a critical yet unresolved challenge in laparoscopy.

Point cloud registration methods have been proposed to align the preoperative organ surface point cloud, extracted from imaging modalities like CT scan, with an intraoperative point cloud, obtained from e.g. stereo vision of the surgical field, by estimating the transformation needed to match corresponding anatomical structures.
A state-of-the-art algorithm in non-rigid partial point cloud registration applied to generic objects is Lepard \cite{li2022lepard}. Their model combines a fully convolutional feature extraction, Transformer-based self cross-attention and differentiable matching to predict point cloud correspondences. 
Despite the excellent performances on synthetic data of deep learning-based point cloud registration, challenges remain in real-world medical applications due to the disparity between dense, noise-free preoperative point clouds and intraoperative point clouds, which are often partial ($<20\%$), noisy \cite{zha2023endosurf}, and deformable, making accurate alignment difficult. To this end, recent methods have been developed to improve registration in challenging surgical environments.
One of the first approaches is \cite{guan2023intraoperative}, which proposes a rigid registration algorithm for liver surfaces in a low overlap scenario. It combines local and global features to estimate the overlap mask used to filter the non-overlapping region of the point cloud. 
LiverMatch \cite{yang2023learning} employs a transformer-based network for matching complete and partial liver point clouds in deformed conditions. Furthermore, the authors propose their synthetic dataset starting from 16 livers in the 3D-IRCADb-01 dataset \cite{soler20103d} and applying deformations and crops to them. 
Zhang et al. \cite{zhang2024point} propose KCR-Net, integrating a Neighborhood Feature Fusion Module (NFFM) for robust keypoint registration, even in low-overlap scenarios. Despite these advancements, challenges remain in handling intraoperative deformations, noise, and complete-to-partial registrations.

All these state-of-the-art methods adopt an agnostic strategy, training on large datasets to generalize to new cases. However, we argue this approach may not be ideal in surgical settings, where error margins are extremely low and optimal outcomes are critical for each patient. To address this limitation, we introduce a novel patient-specific non-rigid point cloud registration method that leverages preoperative information to tailor the registration process for each individual. Our method employs a Transformer encoder-decoder architecture combined with overlap estimation and a dedicated matching module to predict dense correspondences. Additionally, we introduce a novel on-the-fly data generation strategy during training, which includes dynamic patient-specific deformation generation and an innovative visible crop logic. Experimental results on both synthetic and real data confirm that our patient-specific approach significantly outperforms traditional agnostic methods.

 \begin{figure}[t]
    \centering
        \includegraphics[width=\textwidth]{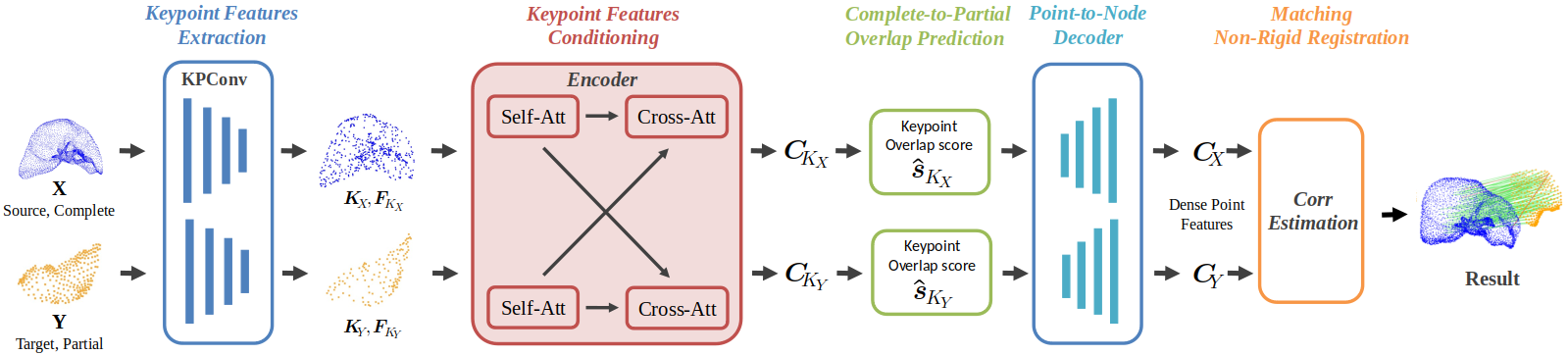}
  \caption{The network uses KPConv to detect keypoints and extract their features, refines these features through self- and cross-attention, and then feeds them into the decoder and matching module to predict dense point correspondences.}
  \label{fig:arch}
\end{figure}
\section{Methods}
The architecture of the proposed registration method is presented in Fig. \ref{fig:arch}. The pipeline consists of five main modules: \textit{Keypoint Features Extraction}, \textit{Keypoint Features Conditioning}, \textit{Complete-to-Partial Overlap Prediction}, \textit{Point-to-Node Decoder} and \textit{Matching Non-Rigid Registration}. First, KPConv is utilized to extract keypoints and their corresponding features from the input point clouds. Specifically, let $\mathbf{X} \in \mathbb{R}^{M \times 3}$ represent the complete point cloud of the organ of interest (e.g., derived from a 3D model obtained via a CT scan), and let $ \mathbf{Y} \in \mathbb{R}^{N \times 3} $ denote a partial point cloud of the same organ (e.g., captured using an endoscopic camera), where $N \ll M$. The extracted features are then processed by the encoder, which employs self-attention and cross-attention mechanisms to enhance feature representation and alignment. The \textit{Complete-to-Partial Overlap Prediction} module predicts the regions of $\mathbf{X}$ and $\mathbf{Y}$ that correspond to the same anatomical area, helping to refine keypoint matching. Keypoint features and overlap scores are extended to the dense representation using a \textit{Point-to-Node Decoder}. Subsequently, the \textit{Matching} module determines correspondences by leveraging the similarity matrix computed between these features and the original 3D coordinates.

\subsubsection{Keypoint Features Extraction Module}
Kernel Point Convolution (KPConv) \cite{thomas2019kpconv} is used for feature extraction. The KPConv backbone converts the input point cloud into a reduced set of keypoints \( \mathbf{K}_X \in \mathbb{R}^{M' \times 3} \) and \( \mathbf{K}_Y \in \mathbb{R}^{N' \times 3} \) through a series of ResNet-like blocks and strided convolutions, effectively downsampling the two clouds to achieve similar densities. In addition to the keypoints, the associated features \( \mathbf{F}_{{K}_X} \in \mathbb{R}^{M' \times D} \) and \( \mathbf{F}_{{K}_Y} \in \mathbb{R}^{N' \times D} \) are also produced.

\subsubsection{Keypoint Features Conditioning (Encoder) Module}
We reduce the dimension of the keypoints features (i.e. $\mathbf{F}_{{K}_X}$ and $\mathbf{F}_{{K}_Y}$) to $d=256$ through a linear projection. Additionally, we apply sinusoidal positional encoding to the keypoint coordinates (i.e. $\mathbf{K}_X$ and $\mathbf{K}_X$) to provide the transformer with an understanding of their spatial relationships. Both the high-level features and positional encodings are fed into the cross-encoder. 
High-level features enable the transformer to interpret and compare semantic content within the point clouds during alignment, while positional encoding supplies spatial context by emphasizing the structure and relative arrangement of points.

Following \cite{yew2022regtr}, each transformer cross-encoder layer consists of three sub-layers: (i) a multi-head self-attention layer that processes each point cloud independently; (ii) a multi-head cross-attention layer that updates the features of one point cloud by incorporating information from the other; and (iii) a position-wise feed-forward network.
\(\mathbf{C}_{{K}_X} \in \mathbb{R}^{M' \times d}\) and \(\quad \mathbf{C}_{{K}_Y} \in \mathbb{R}^{N' \times d}\)
 are the conditioned keypoint features produced by the encoder. 

\subsubsection{Overlap prediction Module}
\label{overlap_pred}

The overlap module leverages conditioned features to compute keypoint overlap scores, denoted as $\mathbf{\hat{s}} = \left[ \mathbf{\hat{s}}_{K_{X}}, \mathbf{\hat{s}}_{K_{Y}}\right]$ using a linear fully connected layer with sigmoid activation, following:
\begin{equation}
\mathbf{\hat{s}} = 
\Bigg\{
\begin{array}{ll}
\mathbf{\hat{s}}_{K_{X}} = 1/(1 + e^{-(\mathbf{C}_{{K}_X} \mathbf{W}_3 + \mathbf{b}_3)}) \\
\mathbf{\hat{s}}_{K_{Y}} = 1/(1 + e^{-(\mathbf{C}_{{K}_Y} \mathbf{W}_3 + \mathbf{b}_3)})
\end{array}
\end{equation}
where $\mathbf{W}_3$ and $\mathbf{b}_3$ are learnable weights and biases parameters. These scores quantify the probability that a keypoint lies within the overlapping region. In our low-overlap scenario—where less than 20\% of \textbf{X} is covered by the target—accurate overlap score prediction is vital \cite{zhao2024deep}, as it directs the model to focus its correspondence search on the relevant region.

\subsubsection{Point-to-Node Decoder}
To establish dense correspondences between the two clouds, we employ a decoder that propagates features and overlap scores across every dense point. Instead of using the conventional KPConv decoder, which relies on k-nearest neighbour search, we adopted a point-to-node grouping strategy similar to \cite{yu2021cofinet}. This approach offers two main advantages: (1) each point is uniquely assigned to one node (i.e. keypoint), ensuring no point is left unassigned, and (2) it inherently adapts to various scales \cite{li2019usip}. After grouping, we obtain dense point features (\(\mathbf{C}_{X} \in \mathbb{R}^{M \times d}\), \(\quad \mathbf{C}_{Y} \in \mathbb{R}^{N \times d}\)) and overlap scores (\(\mathbf{\hat{s}}_{X}, \mathbf{\hat{s}}_{Y})\). The features are concatenated with the point's 3D coordinates (\(\mathbf{C}_X^{*} \in \mathbb{R}^{M \times d+3}\), \(\quad \mathbf{C}_Y^{*} \in \mathbb{R}^{N \times d+3}\)), and this combined vector is fed into a MLP to estimate the final coordinates for \textbf{X}. Our MLP first projects the concatenated input from a $d+3$ dimensional space into a $d$-dimensional representation, applies non-linear activations, and ultimately outputs a 3-dimensional vector corresponding to the deformed coordinates (\(\mathbf{\hat{X}},  \mathbf{\hat{Y}}\)). 
Specifically: $\mathbf{\hat{X}} = \text{ReLU}(\mathbf{C}_{X}^{*} \mathbf{W}_1 + \mathbf{b}_1) \mathbf{W}_2 + \mathbf{b}_2$, where \(\mathbf{W}_1, \mathbf{W}_2\) and \(\mathbf{b}_1, \mathbf{b}_2\) are learnable weights and biases, respectively. At this stage, the model attempts an initial non-rigid registration. Although the deformed coordinates it produces are rough, they provide a valuable signal for the chamfer loss to effectively guide the training process. Separately, the overlap scores are refined through a linear layer.

\subsubsection{Matching and Physics-based Non-rigid Registration}
Within the Matching module, we begin by refining the $\mathbf{C}_{X}$ and $\mathbf{C}_{Y}$ using a linear layer. Then, we compute a similarity matrix and convert it into a confidence matrix \textbf{M} using a dual-softmax operation. Finally, following \cite{yang2023learning}, matches $\mathbf{m}$ are selected from \textbf{M} based on a thresholded Mutual Nearest Neighbor criterion.

Once the matching is obtained, we can formulate the non-rigid registration problem as an energy minimization: $\underset{\delta\mathbf{X}}{\min} \frac{1}{2} \delta\mathbf{X}^T \mathbf{S} \delta\mathbf{X} + k \| \mathbf{Y}_\mathbf{m} - \mathbf{X}_\mathbf{m} \|^2$, 
where $\delta\mathbf{X}$ is the displacement field to be estimated, $\mathbf{S}$ is the stiffness matrix, which encodes the biomechanical properties of the organ, $k$ is a scalar stiffness parameter that converts displacement to external forces.

The first term $\frac{1}{2} \delta\mathbf{X}^T \mathbf{S} \delta\mathbf{X}$ represents the internal elastic energy, ensuring that deformations follow biomechanical constraints, in particular in non-visible parts of the organ. The second term, $k \| \mathbf{Y}_\mathbf{m} - \mathbf{X}_\mathbf{m} \|^2$, enforces alignment with the observed correspondences. Since this formulation is static (no acceleration), the system does not include mass or damping terms. This system is solved using the Conjugate Gradient method.

\subsubsection{Optimization Losses}
Our method employs a weighted sum of three losses: (i) a matching loss (ML) \cite{lin2017focal} to supervise the confidence matrix \textbf{M} with the ground truth matches; (ii) a chamfer loss (CL) weighted on the overlap score, defined as:  $CL(\mathbf{\hat{X}}, \mathbf{Y}, \mathbf{\hat{s}}_{X}) = \frac{1}{|\mathbf{\hat{X}}|} \sum_{\hat{x} \in \hat{X}} \hat{s}_x \min_{y \in Y} \| \hat{x} - y \|^p $. 
This loss is not focused on finding exact 1-1 correspondences but rather a global geometrical alignment. Finally, (iii) an overlap loss (OL) similar to \cite{yew2022regtr} designed to optimize the overlap scores.

\subsubsection{Patient-specific Training Dataset Generation}
\label{data_gen}
Generating a patient-specific training dataset involves creating pairs of complete and partial 3D point clouds from pre-operative CT scans, incorporating both rigid and non-rigid transformations. 
We employ the As-Rigid-As-Possible (ARAP) algorithm \cite{sorkine2007rigid} to generate two types of deformation: (i) compression deformation, which simulates the effect of \( CO_2 \) insufflation by displacing control points along their normal vectors with randomly assigned magnitudes in the range \([0, 0.1]\); and (ii) lobe deformations, introduced by applying random displacements to control points within the lobe region, with magnitudes ranging from \([0, 0.25]\).

To realistically generate partial point clouds similar to an endoscopic scene, we generate a dummy camera position that points toward the visible surface of the organ. We sample random spherical coordinates while constraining the polar and azimuthal angles to a realistic range for intraoperative settings. Using the camera direction and the surface normals of the organ, we compute the dot product to assess visibility. Only 5\% of points closest to the camera, with a dot product below $80^\circ$, are retained. 

Following the criteria above, we apply deformation, and cropping is applied randomly to constitute $\mathbf{Y}$, while a random rigid transformation is applied to $\mathbf{X}$. In addition, both $\mathbf{X}$ and $\mathbf{Y}$ are shuffled. 
This pipeline is executed on the fly for each intraoperative cloud, with a processing time of 0.5 seconds per sample. Since random parameters govern deformations and cropping, the model is exposed to a continuously varying dataset during training, eliminating the need for a pre-computed offline dataset. Conversely, validation is performed using deterministic seeds to ensure consistency. The test set follows the same generation process as validation but with a different seed.

 \begin{figure}[t]
    \centering
        \includegraphics[width=\textwidth]{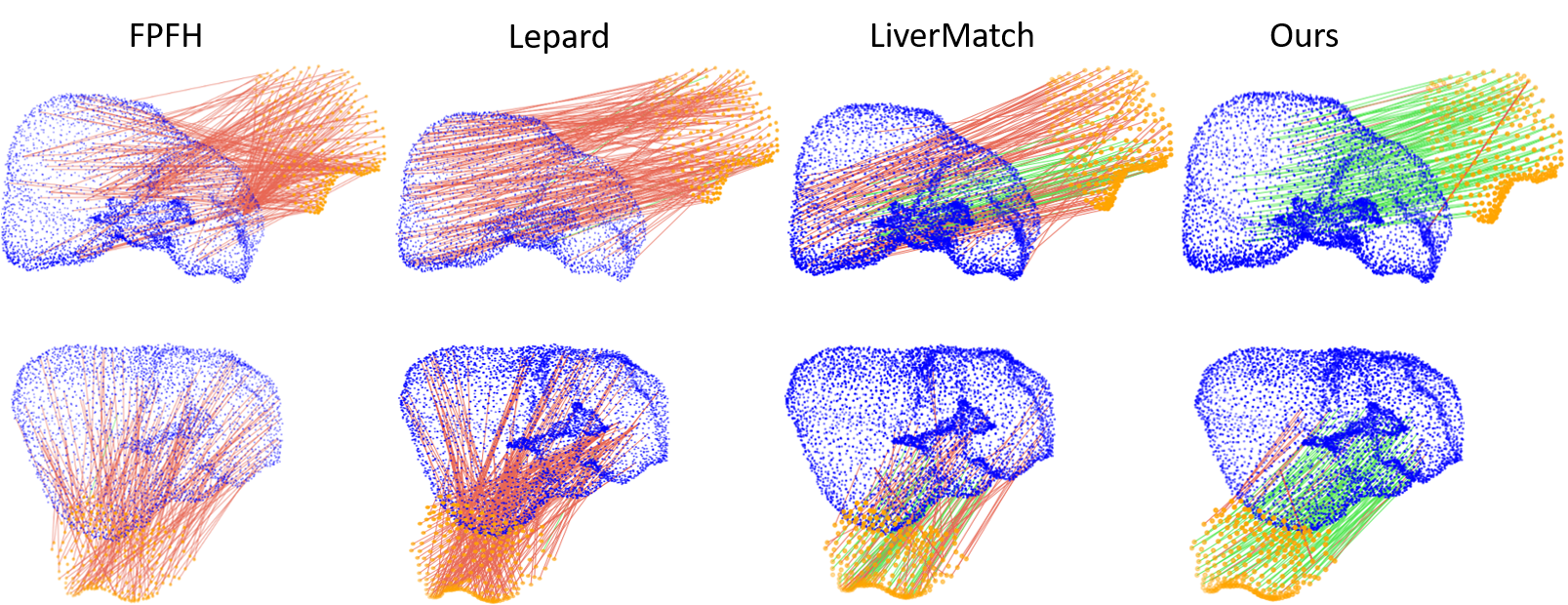}
  \caption{Qualitative results of predicted matches. Blue points denote the preoperative point cloud, while orange points represent the partial and deformed intraoperative cloud. Red lines indicate incorrect matches, and green lines highlight exact matches relative to the ground truth.}
  \label{fig:matchig_res_ircad}
\end{figure}

\section{Results}
\subsubsection{Matching Performances} To evaluate the matching performance of our model, we employed “liver 1” from the 3D-IRCADb-01 dataset and generated a test set comprising 50 examples, following the procedure described in Section \ref{data_gen}. We compared our approach with three alternative algorithms: (i) FPFH – where FPFH features \cite{rusu2009fast} are extracted from both complete and partial point clouds, and correspondences are established by matching each target point to its nearest source point in the feature space using Euclidean distance; (ii) Lepard \cite{li2022lepard} – by leveraging their pre-trained model on the 4DMatch dataset, a collection of dynamic, real-world, partially overlapping point cloud pairs; and (iii) LiverMatch \cite{yang2023learning} – utilizing the pre-trained model derived from liver data within the 3D-IRCADb-01 dataset. Notably, “liver 1” is absent from the LiverMatch training set, as it was also used for test set generation in their work, we call IRCAD-Liver1.
We evaluated performance using the Matching Score (MS)—defined as the number of correctly predicted matches relative to the total available matches—and the Inlier Ratio (IR), which represents the proportion of correct matches among all predictions. Additionally, we report the absolute number of exact matches predicted (\#MP) by each algorithm. Table \ref{tab:results} demonstrates that our algorithm significantly outperforms its competitors in identifying exact matches. Furthermore, we evaluated our algorithm under various loss configurations. Our results indicate that optimal performance is achieved when all the three losses are active (\textbf{\textit{Full}}), confirming the synergistic relationship between overlap score estimation and correspondence estimation. Figure \ref{fig:matchig_res_ircad} visually compares the quality of the matches predicted by each method.



\renewcommand{\arraystretch}{0.5}
\setlength{\tabcolsep}{8pt}
\begin{table}[t]
\centering
\caption{Matching Metrics}
\label{tab:results}
\resizebox{0.7\textwidth}{!}
{
\begin{tabular}{l|ccccc}
\toprule
     Methods  & MS (\%) & IR (\%) & Avg. \#MP\\
\midrule
\rowcolor{gray!10}
\textit{FPFH}  & $0.03 \pm 0.11$    & $0.03 \pm 0.11$ & $<1$ \\
\textit{Leopard} & $1.6 \pm 1.5$      & $2.4 \pm 2.2$ & $4 \pm 3.75$ \\
\rowcolor{gray!10}
\textit{LiverMatch} &   $13 \pm 9$         & $23 \pm 13$ & $39 \pm 27$\\
\midrule
\textit{Ours - ML + CL} &  $36 \pm 8$	& $90 \pm 4$  & $90 \pm 20$\\
\rowcolor{gray!10}
\textbf{\textit{Ours - Full}} &   $45 \pm 7$      & $92 \pm 3$ & $112.5 \pm 17.5$ \\  
\bottomrule
\end{tabular}
}
\end{table}


\subsubsection{Non-rigid Registration}
\begin{figure}[h!]
\subfloat[Matching]{\includegraphics[clip, trim=200 350 200 100, width=0.24\linewidth]{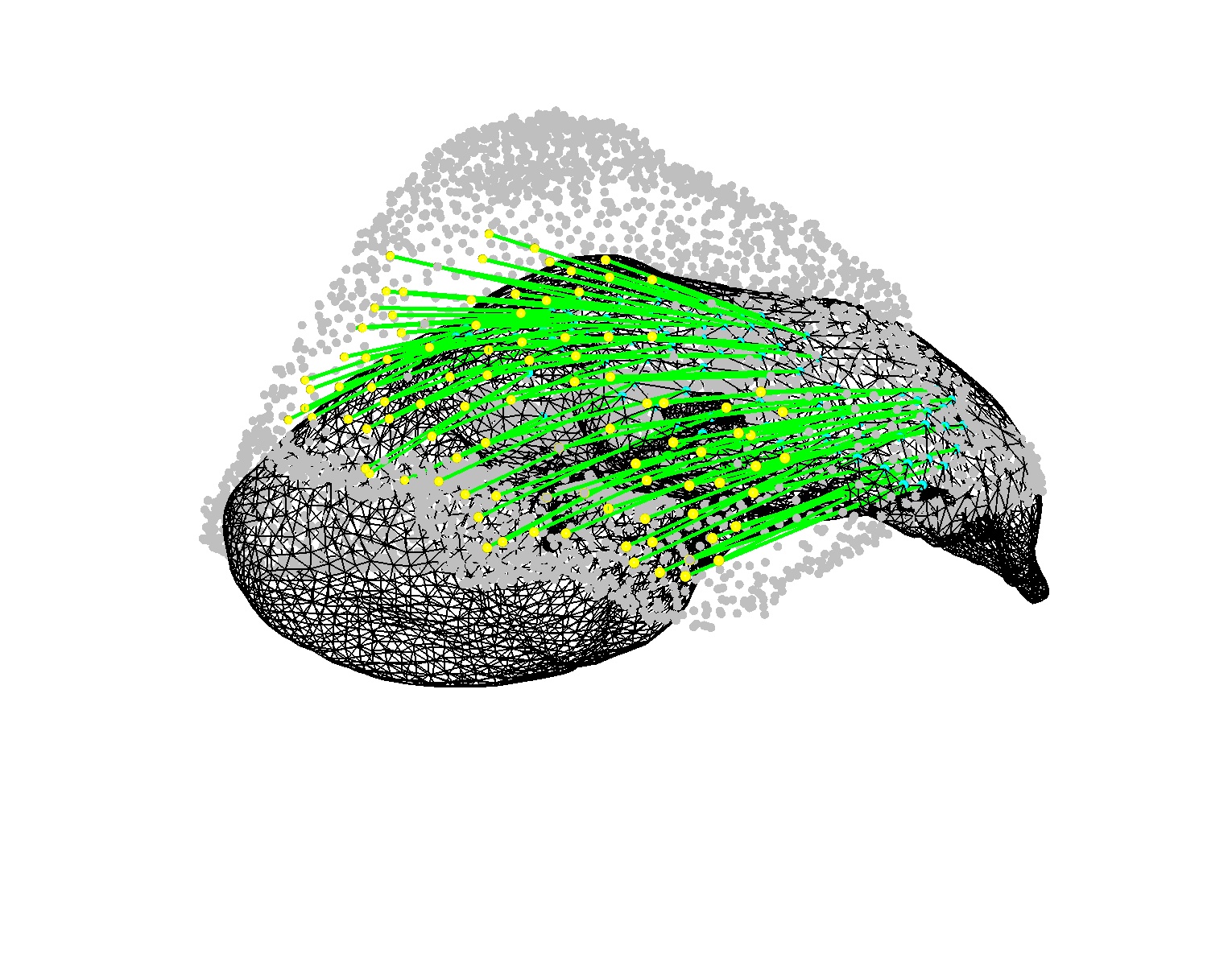}} 
\hfill
\rule{0.5pt}{1.5cm}
\hfill
\subfloat[t=0]{\includegraphics[clip, trim=200 350 200 100, width=0.24\linewidth]{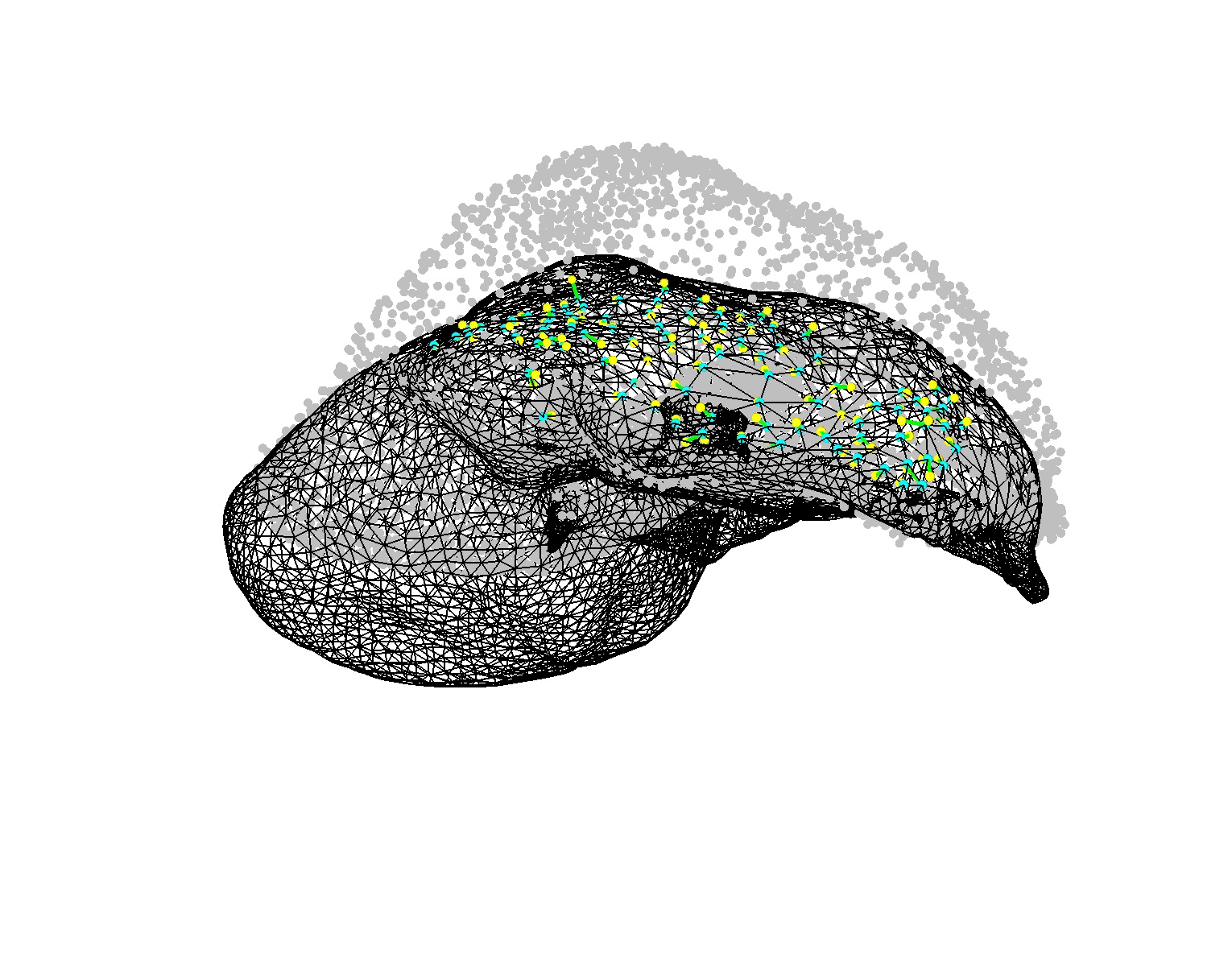}}
\subfloat[t=T/2]{\includegraphics[clip, trim=200 350 200 100, width=0.24\linewidth]{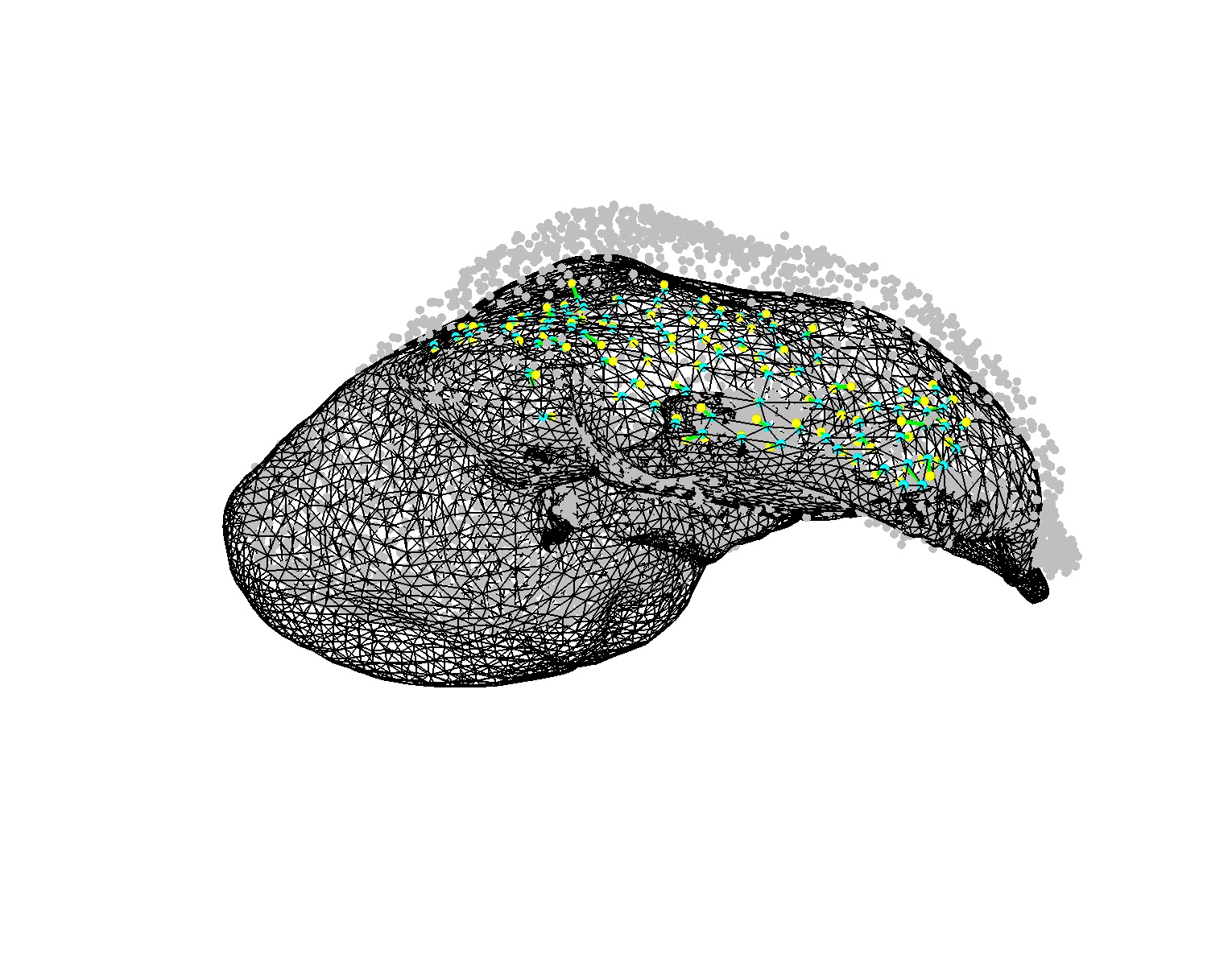}}
\subfloat[t=T]{\includegraphics[clip, trim=200 350 200 100, width=0.24\linewidth]{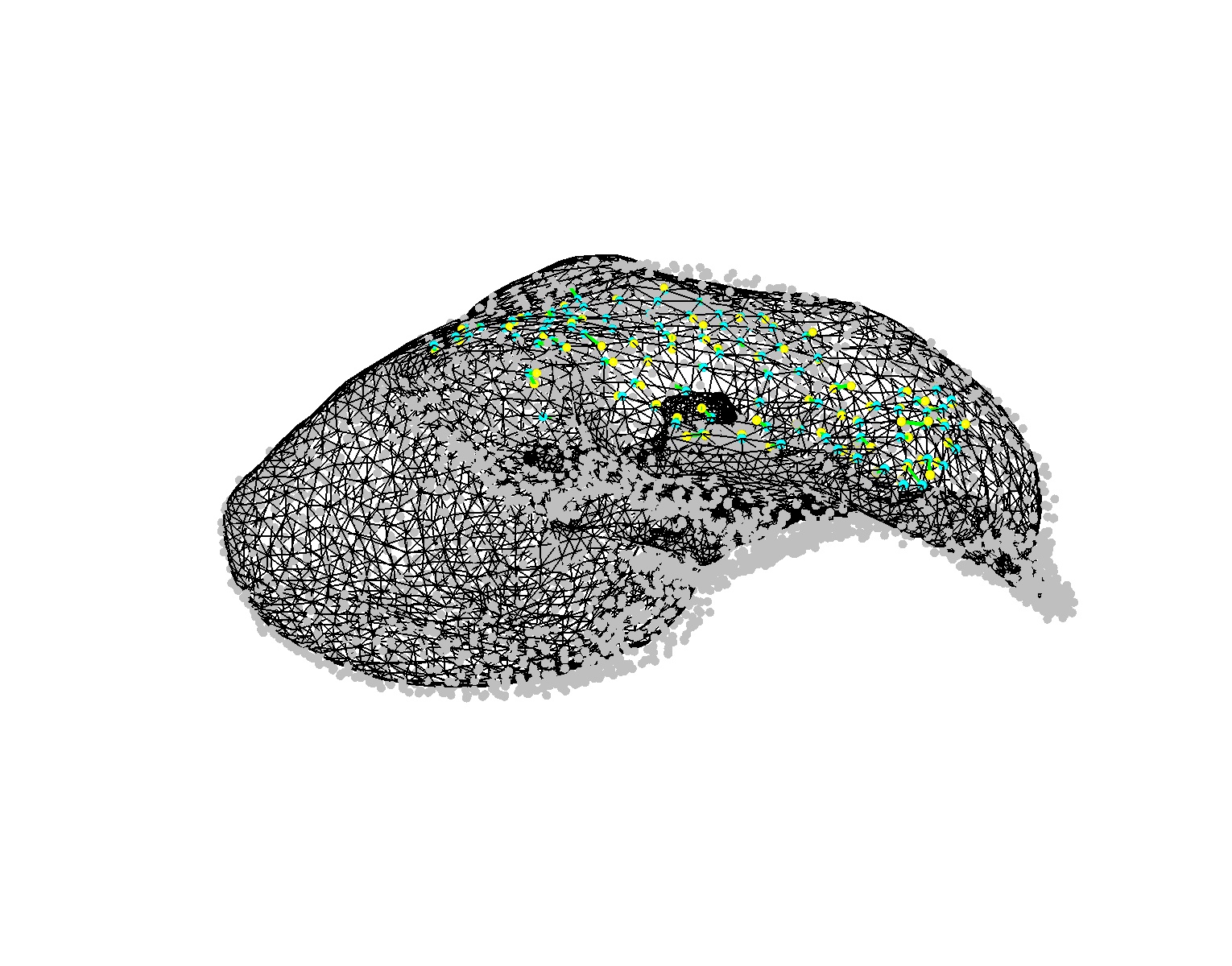}}
\\ 
\subfloat[FEM]{\includegraphics[clip, trim=200 350 200 100, width=0.24\linewidth]{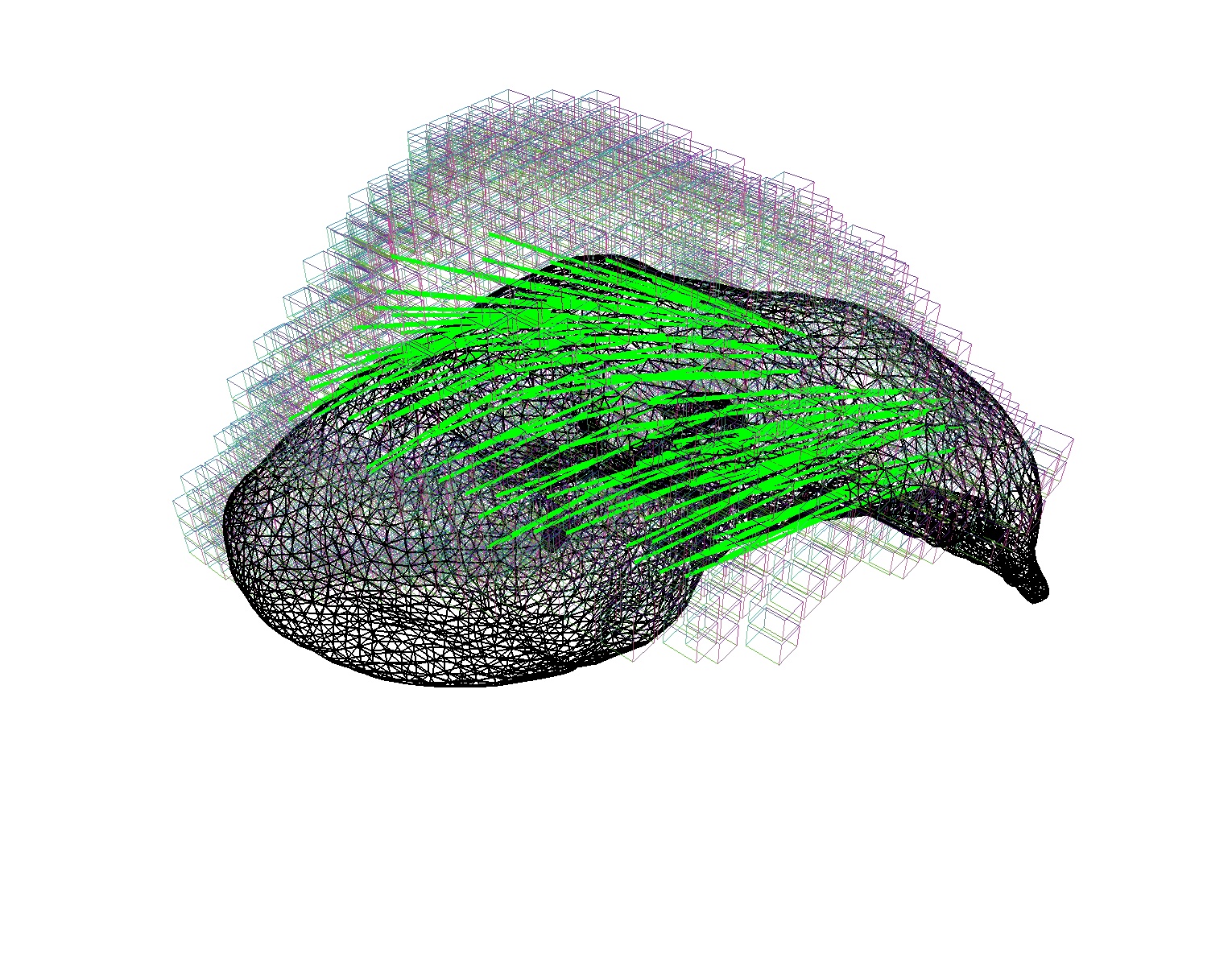}}
\hfill
\rule{0.5pt}{1.5cm}
\hfill
\subfloat[t=T (View 2)]{\includegraphics[clip, trim=150 50 250 300, width=0.24\linewidth]{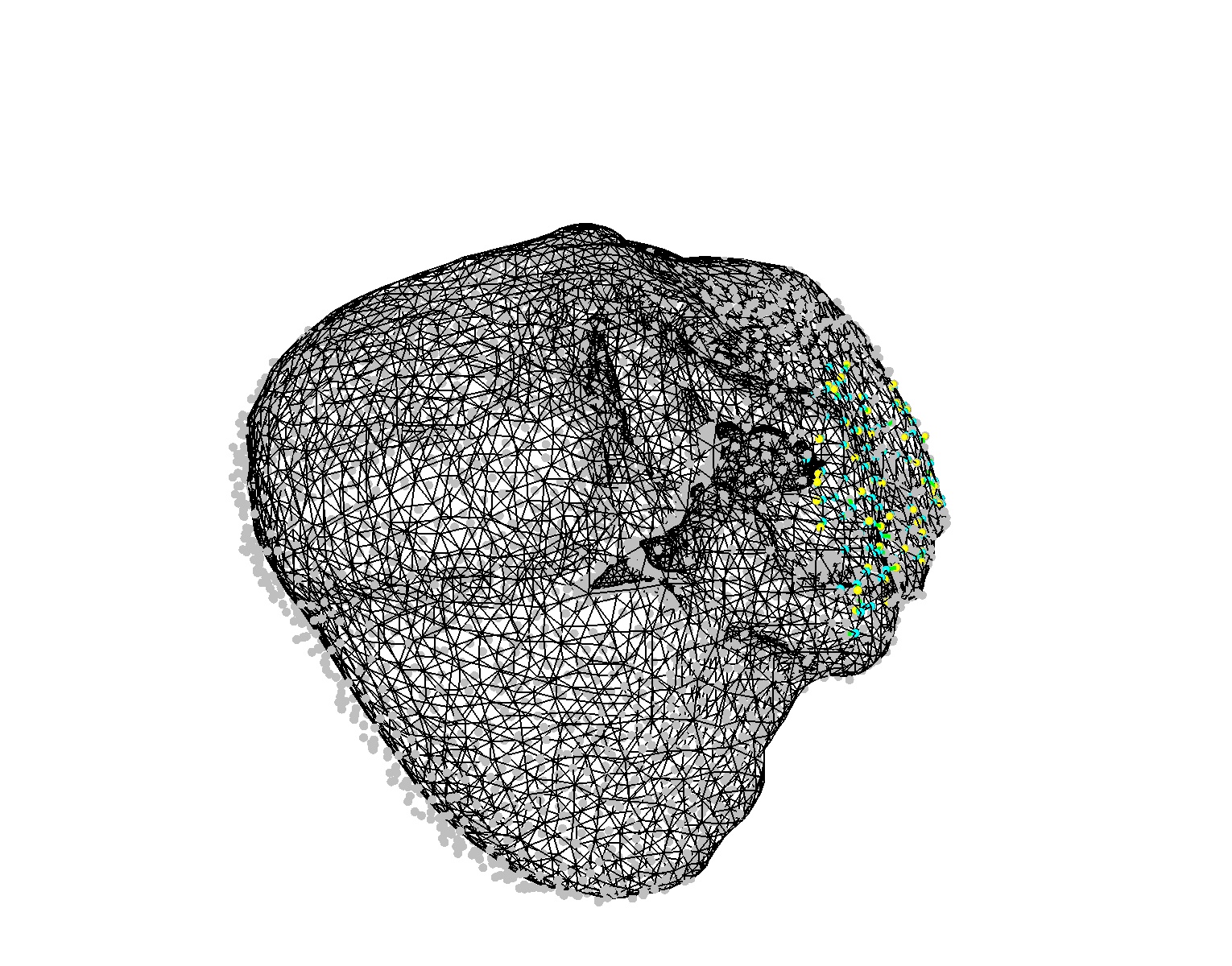}}
\subfloat[t=T (View 3)]{\includegraphics[clip, trim=290 200 200 280, width=0.24\linewidth]{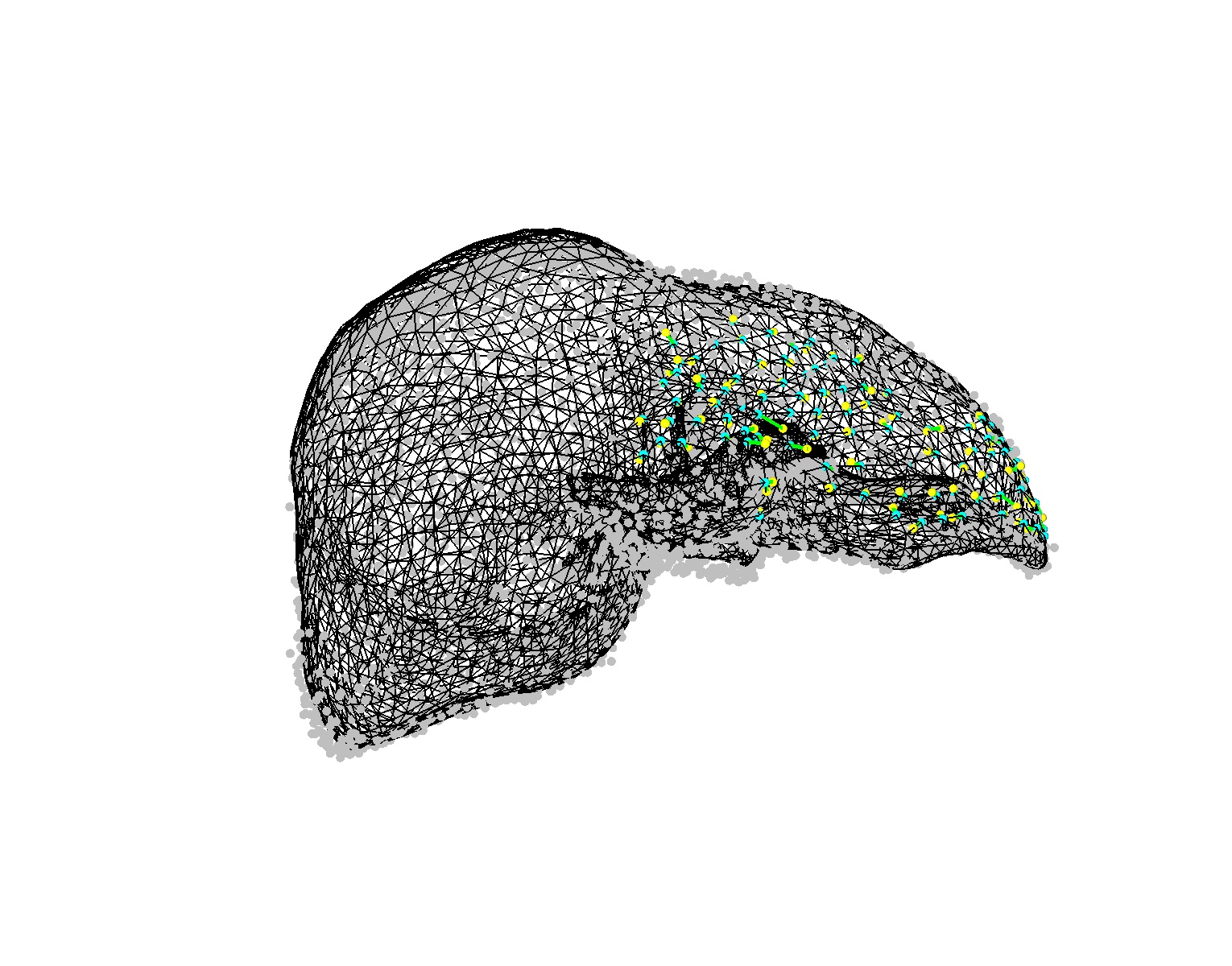}}
\hfill
\subfloat[t=T (View 4)]{\includegraphics[clip, trim=100 100 100 100, width=0.22\linewidth]{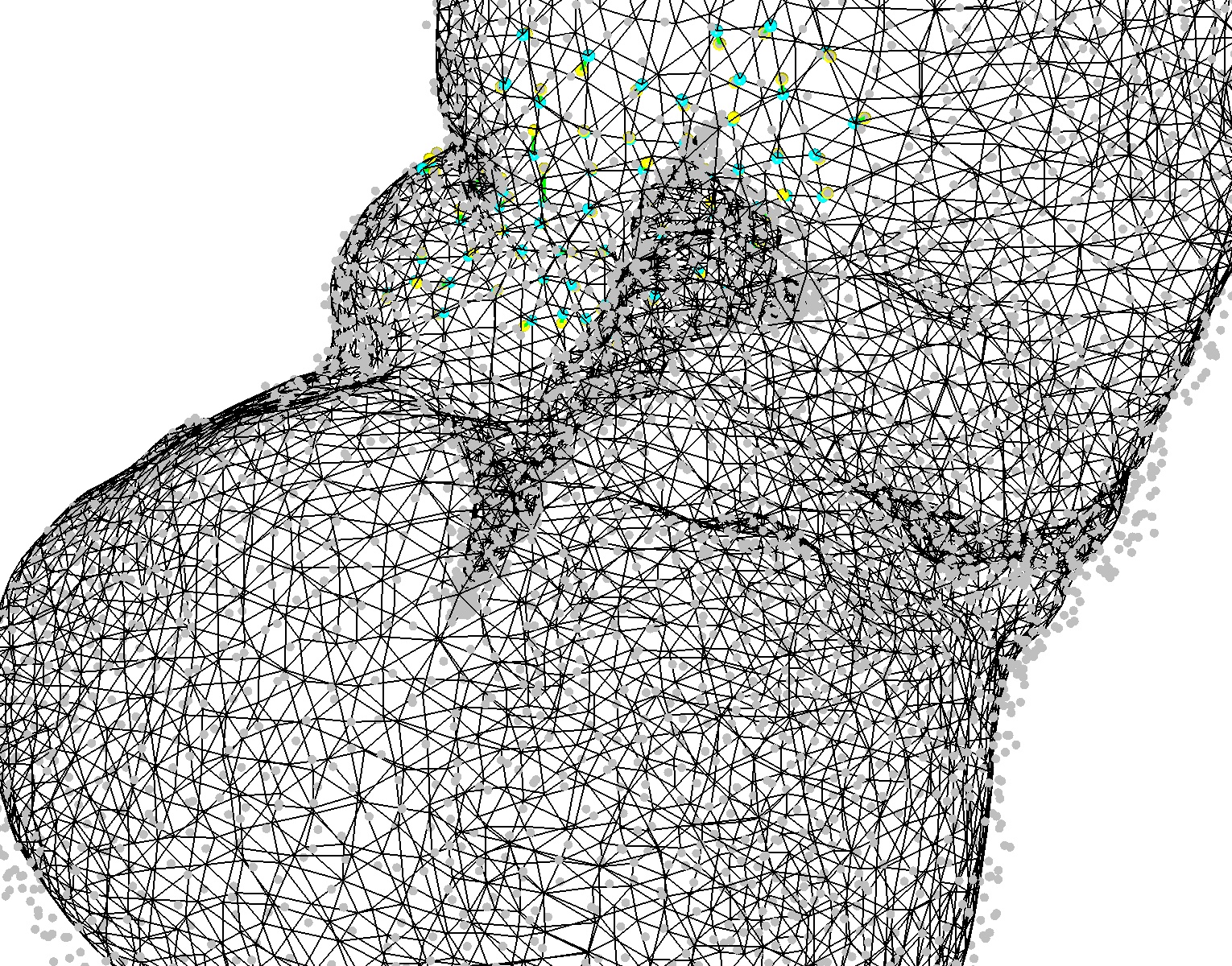}}

\caption{Non-rigid registration results on IRCAD-Liver1: Source point cloud is represented with gray points while target point clouds are represented in black wireframe. Green lines represent the matching correspondences and FEM models are represented with grid cells. Registered point clouds are presented at t=0, t=T/2 and t=T are shown in 4 different views.}
\label{fig:ircad}
\end{figure}

We present results on non-rigid registration using biomechanical modeling.
We use the framework SOFA \cite{sofa} to generate a Finite Element model from the liver mesh and generated around 12000 tetrahedra elements. 
We use a Young's modulus of 1.5 KPa and a Poisson ratio of 0.45.
We compute Target Registration Errors (TREs) as the point-to-point Euclidean distance between vertices of the non-visible part of the preoperative and intraoperative meshes. We also report Fidicual Registration Errors (FREs).
Results shown in Table \ref{tab:registration} using 10 random deformations demonstrate that we obtain very low TREs with an average of $4.82 \pm 3.33$ mm. FREs are also very low, $1.68 \pm 1.11$ mm, suggesting few matching outliers and adequate registration. Moreover, our method outperforms LiverMatch, which exhibits a high number of mismatches, resulting in a TRE of \(17.36 \pm 13.68\)mm and an FRE of \(28.76 \pm 21.30\)mm. Visualization of the non-rigid registration and FEM model is provided in Figure \ref{fig:ircad}.

\captionsetup[subfigure]{labelformat=empty}

\begin{figure}[h!]
\subfloat{\includegraphics[clip, trim=0 0 600 0, width=0.12\linewidth]{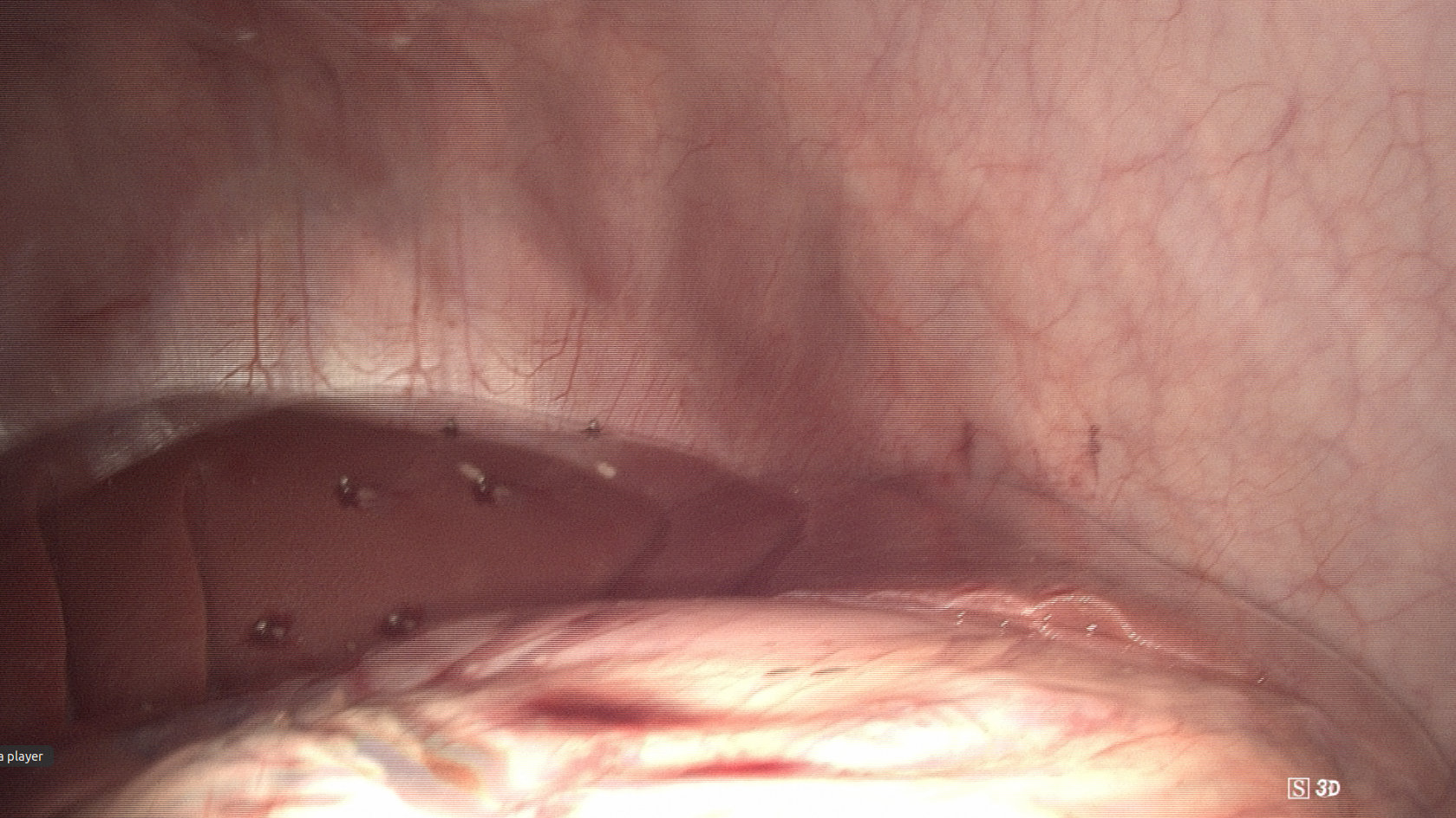}}
\hfill
\subfloat{\includegraphics[clip, trim=300 300 300 300, width=0.18\linewidth]{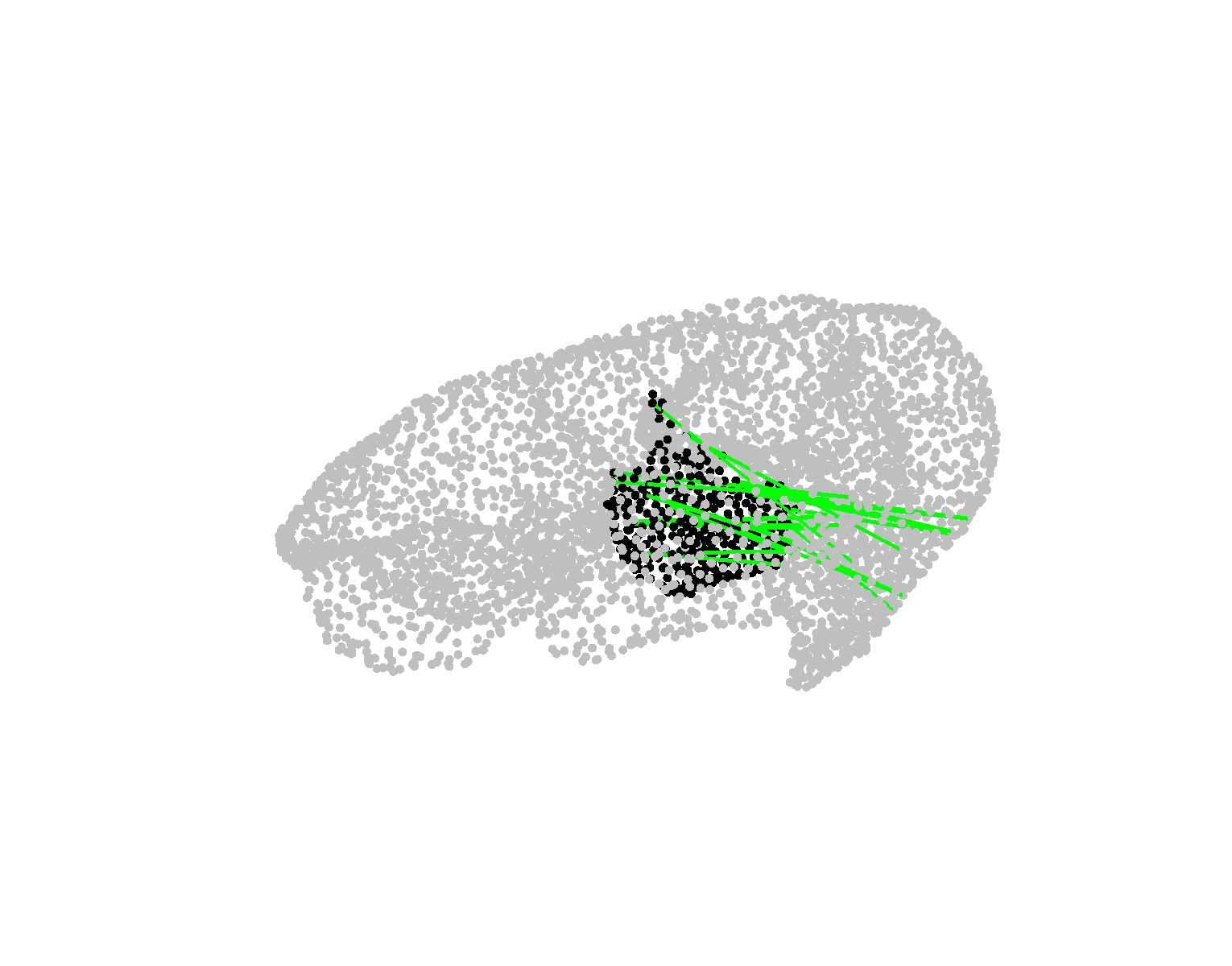}}
\hfill
\subfloat{\includegraphics[clip, trim=300 300 300 300, width=0.18\linewidth]{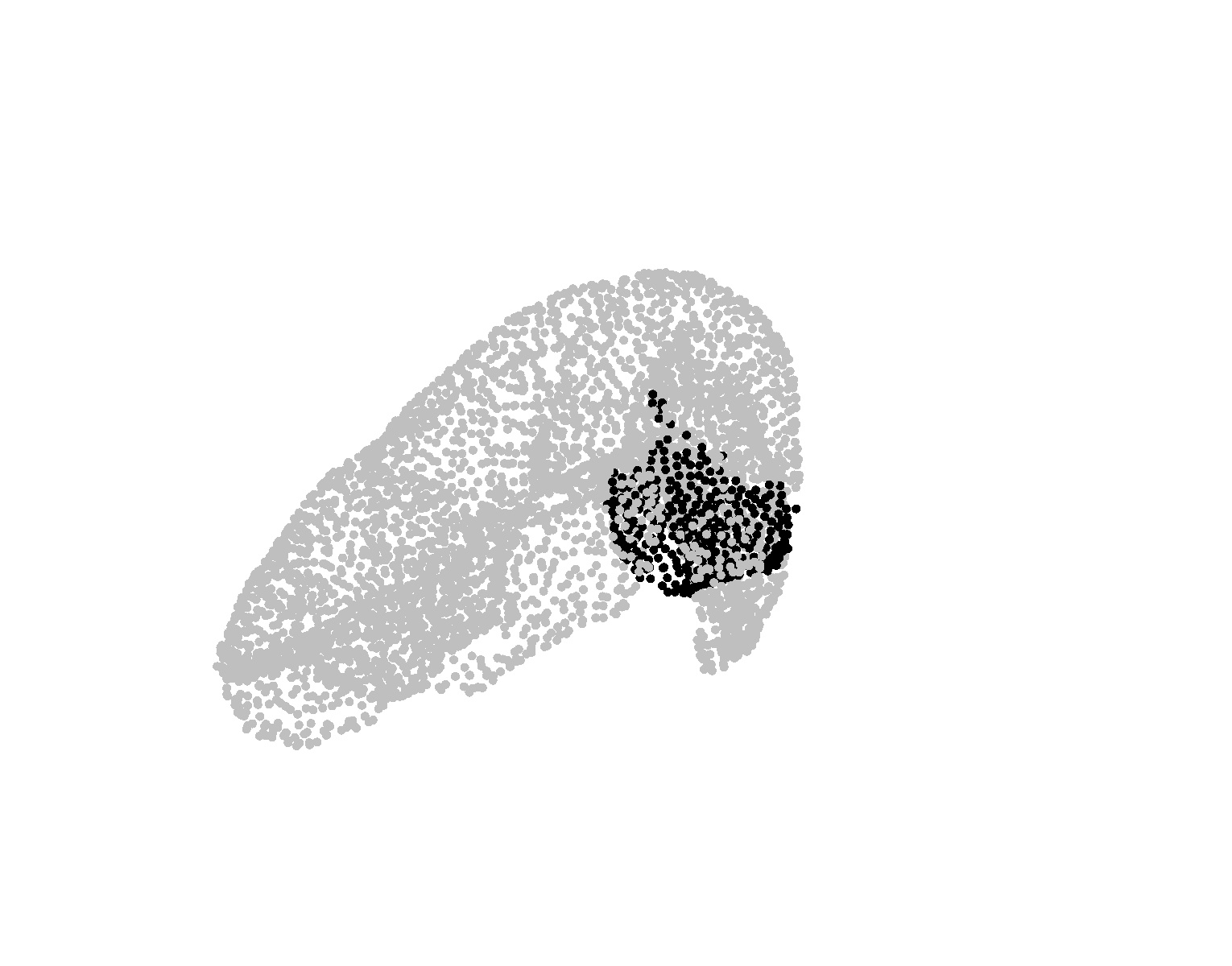}}
\hfill
\subfloat{\includegraphics[clip, trim=0 0 600 0, width=0.12\linewidth]{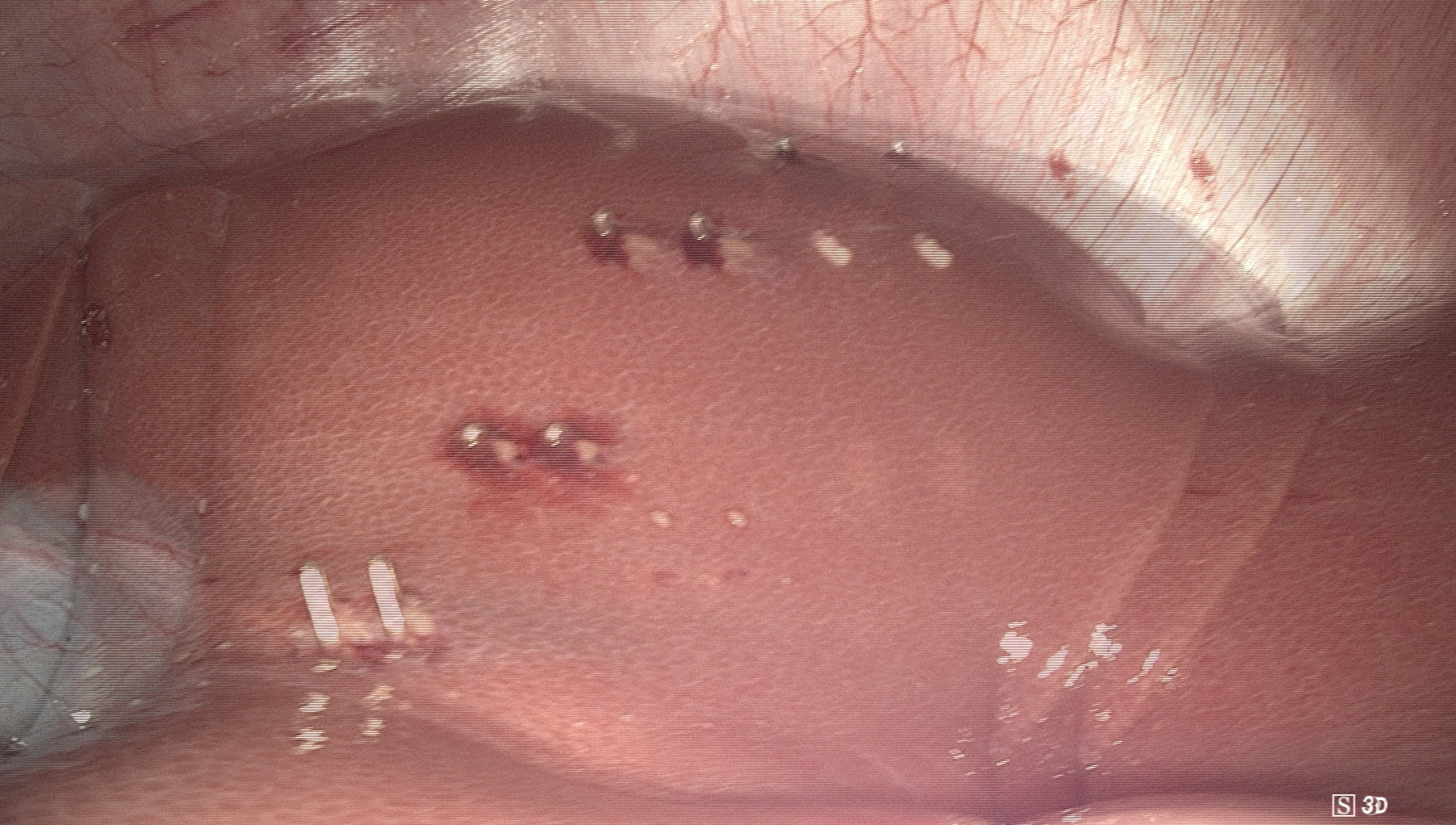}}
\hfill
\subfloat{\includegraphics[clip, trim=300 300 300 300, width=0.18\linewidth]{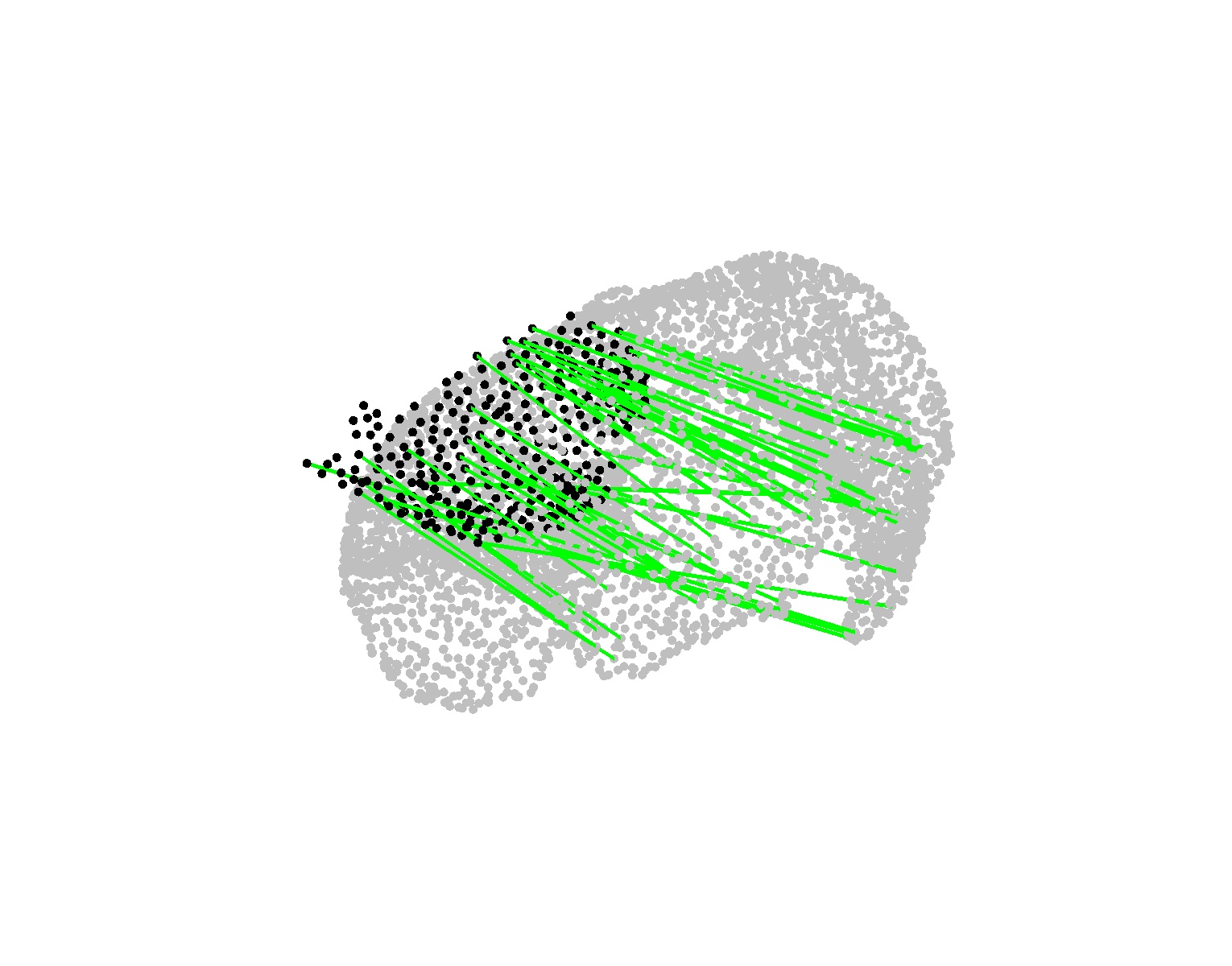}}
\hfill
\subfloat{\includegraphics[clip, trim=150 300 550 300, width=0.18\linewidth]{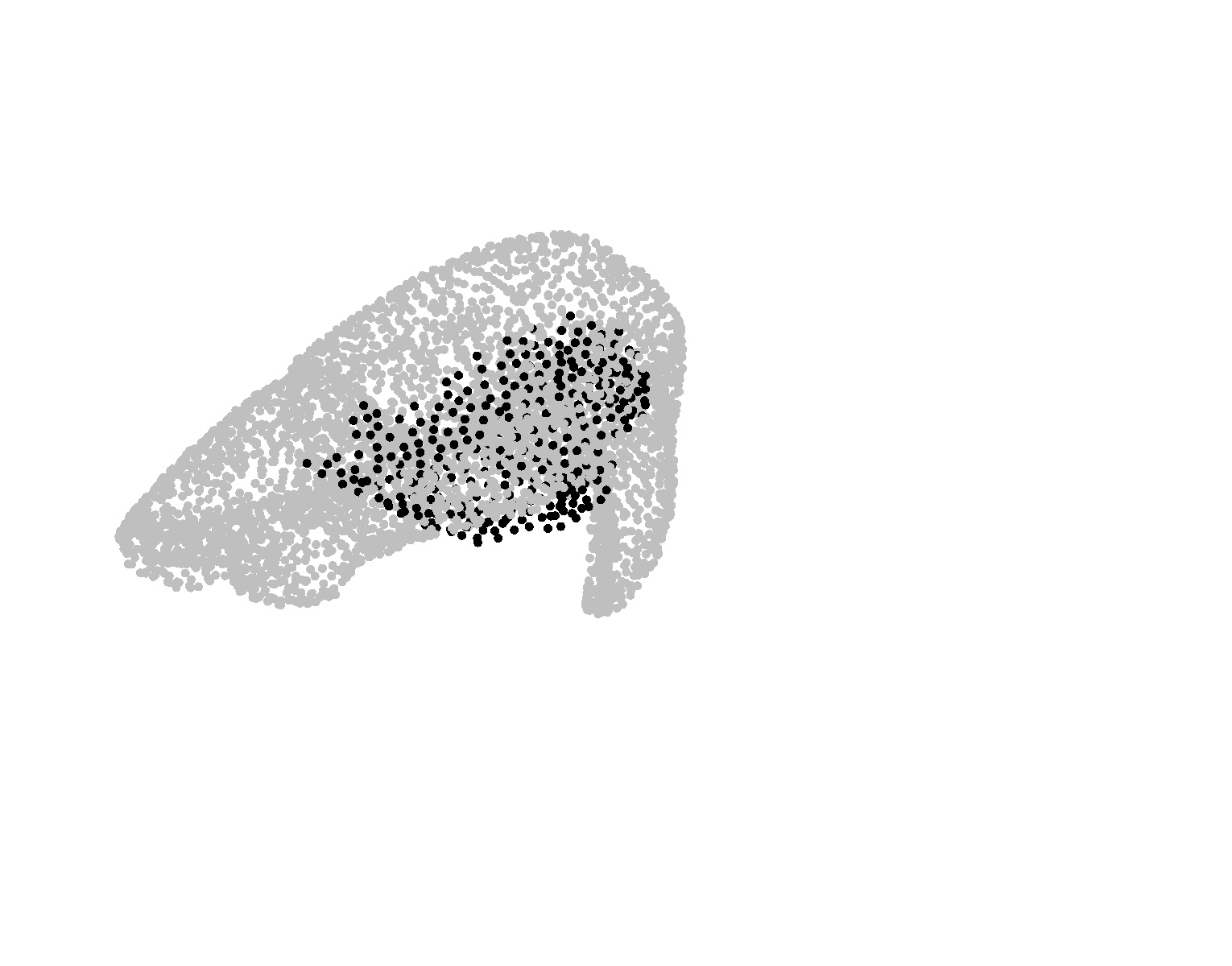}}
\\
\subfloat{\includegraphics[clip, trim=300 0 600 0, width=0.12\linewidth]{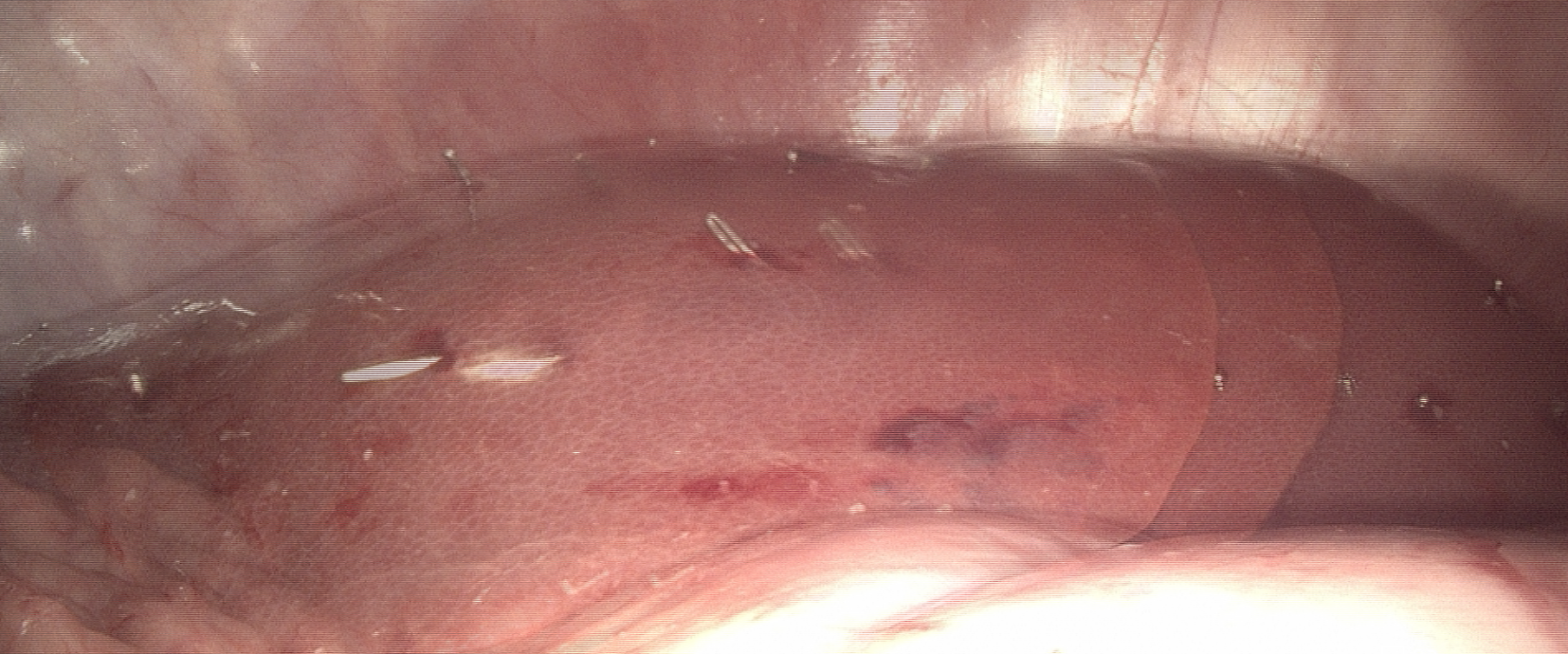}}
\hfill
\subfloat{\includegraphics[clip, trim=300 300 300 300, width=0.18\linewidth]{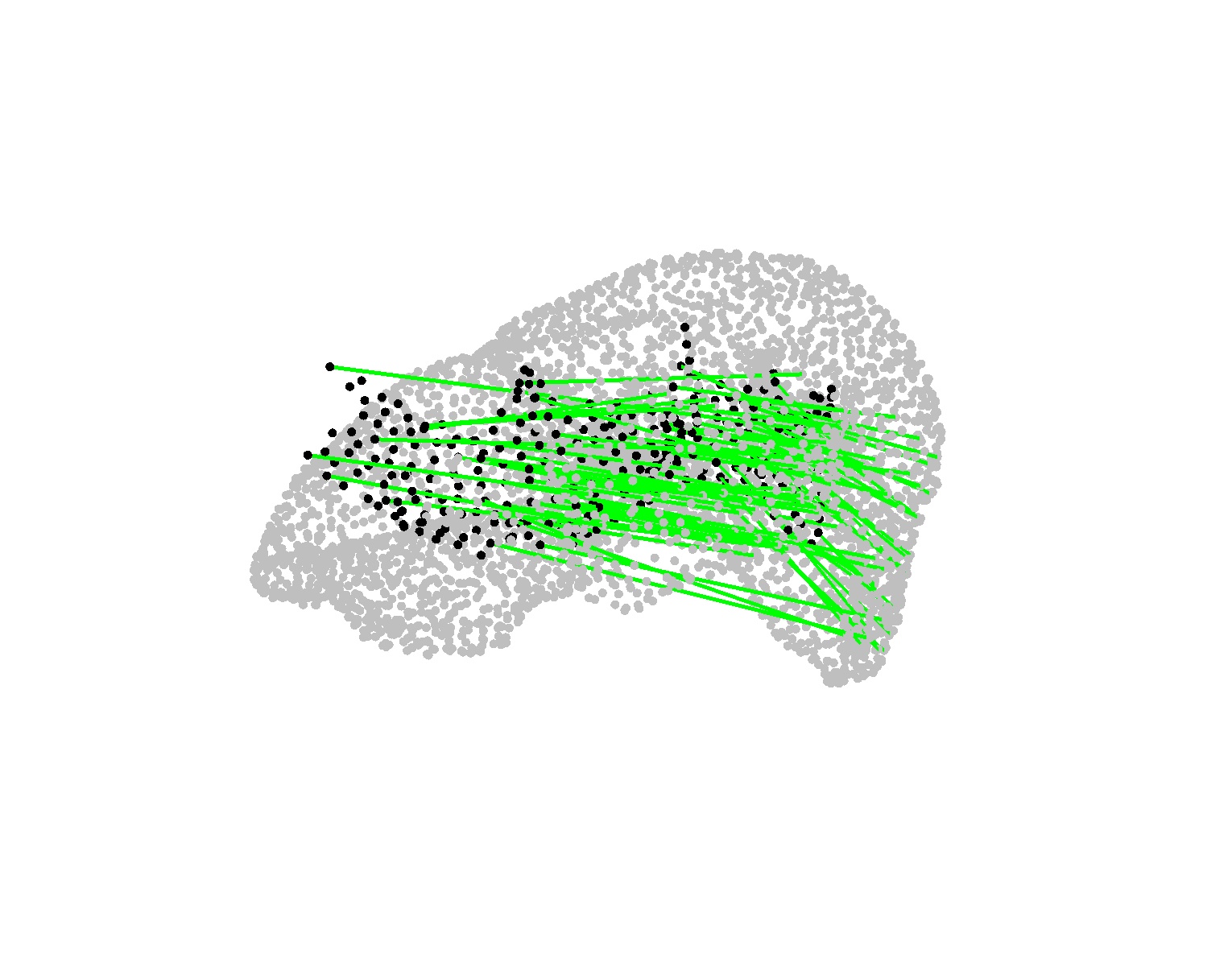}}
\hfill
\subfloat{\includegraphics[clip, trim=200 300 400 200, width=0.18\linewidth]{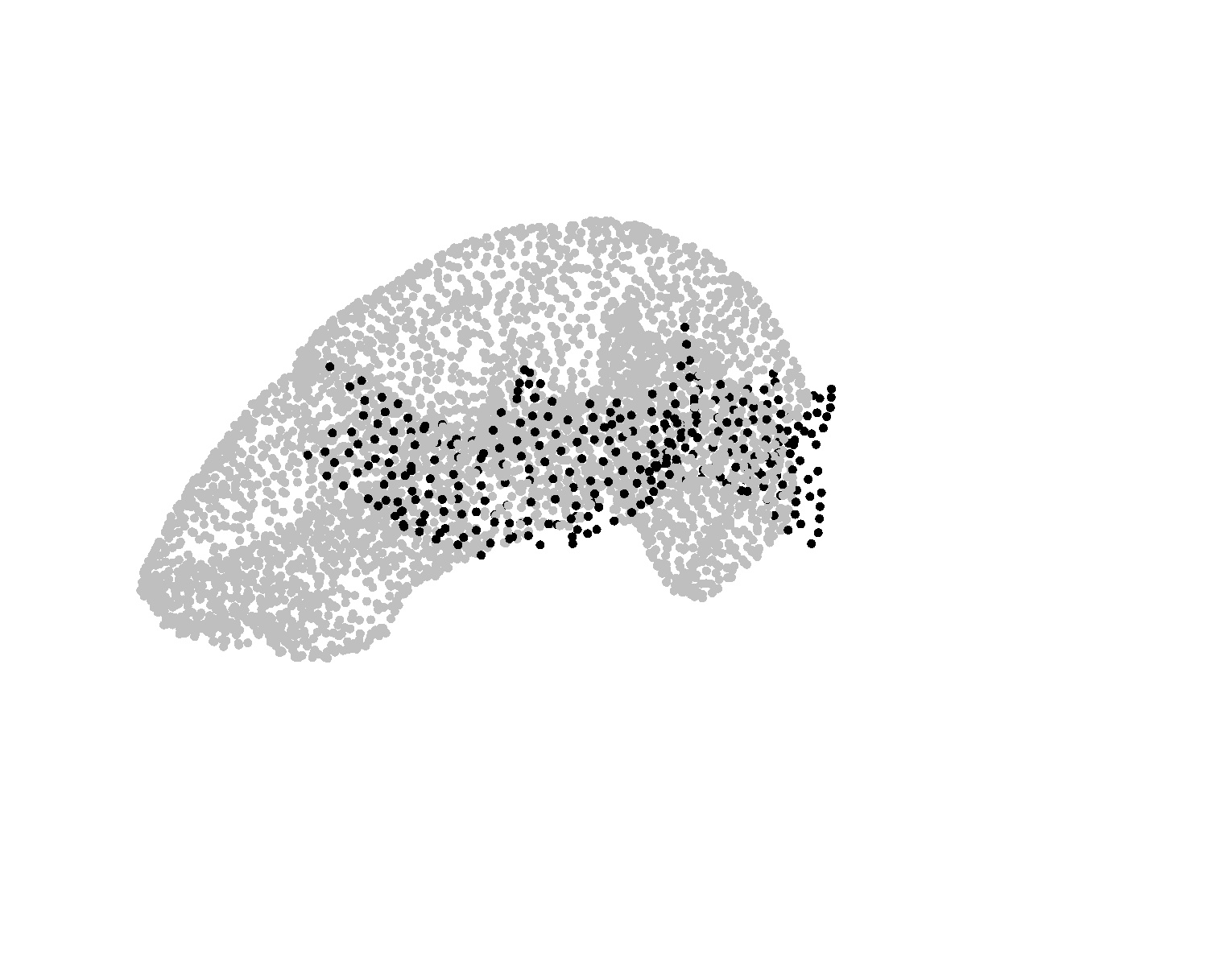}}
\hfill
\subfloat{\includegraphics[clip, trim=300 0 300 0, width=0.12\linewidth]{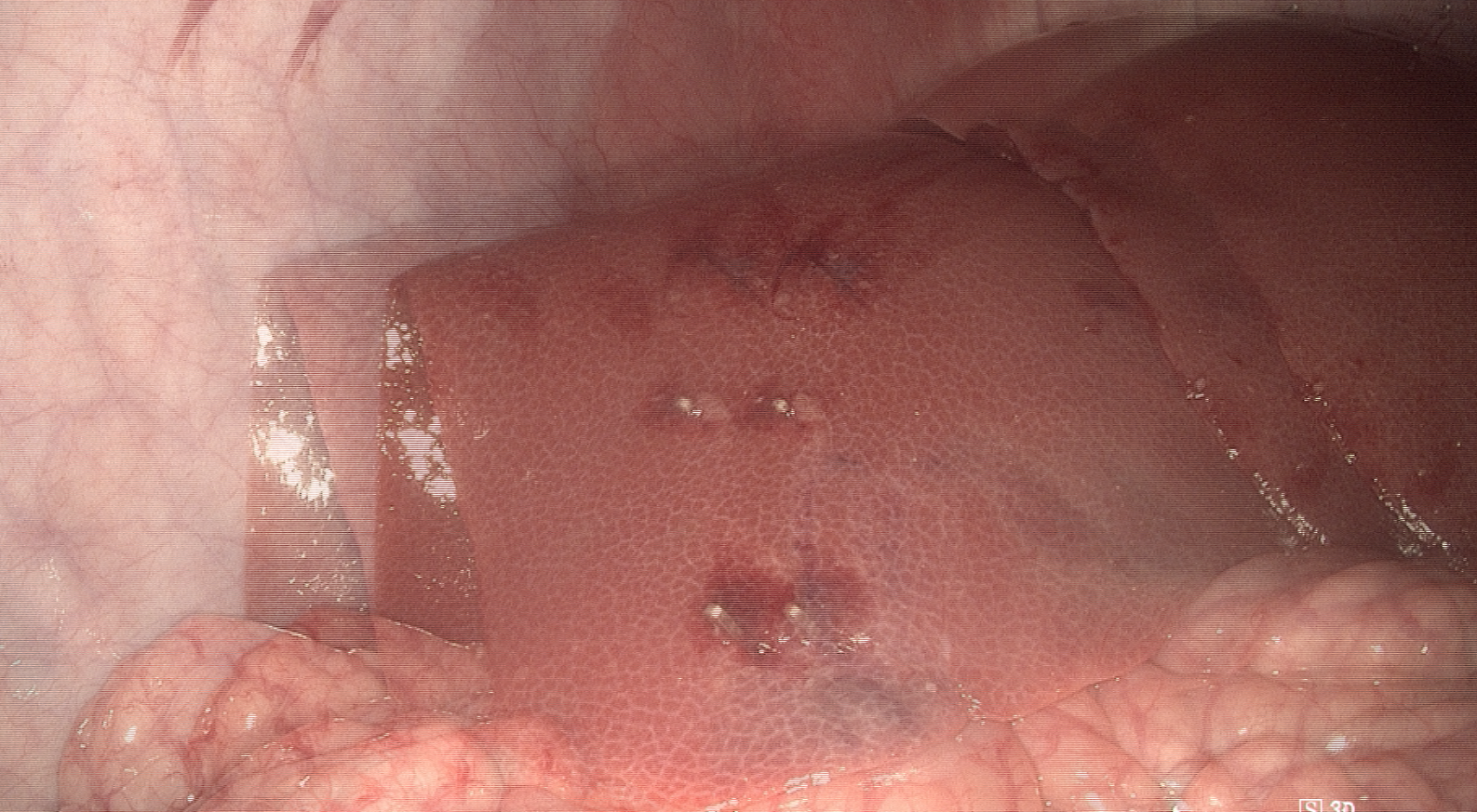}}
\hfill
\subfloat{\includegraphics[clip, trim=500 300 100 300, width=0.18\linewidth]{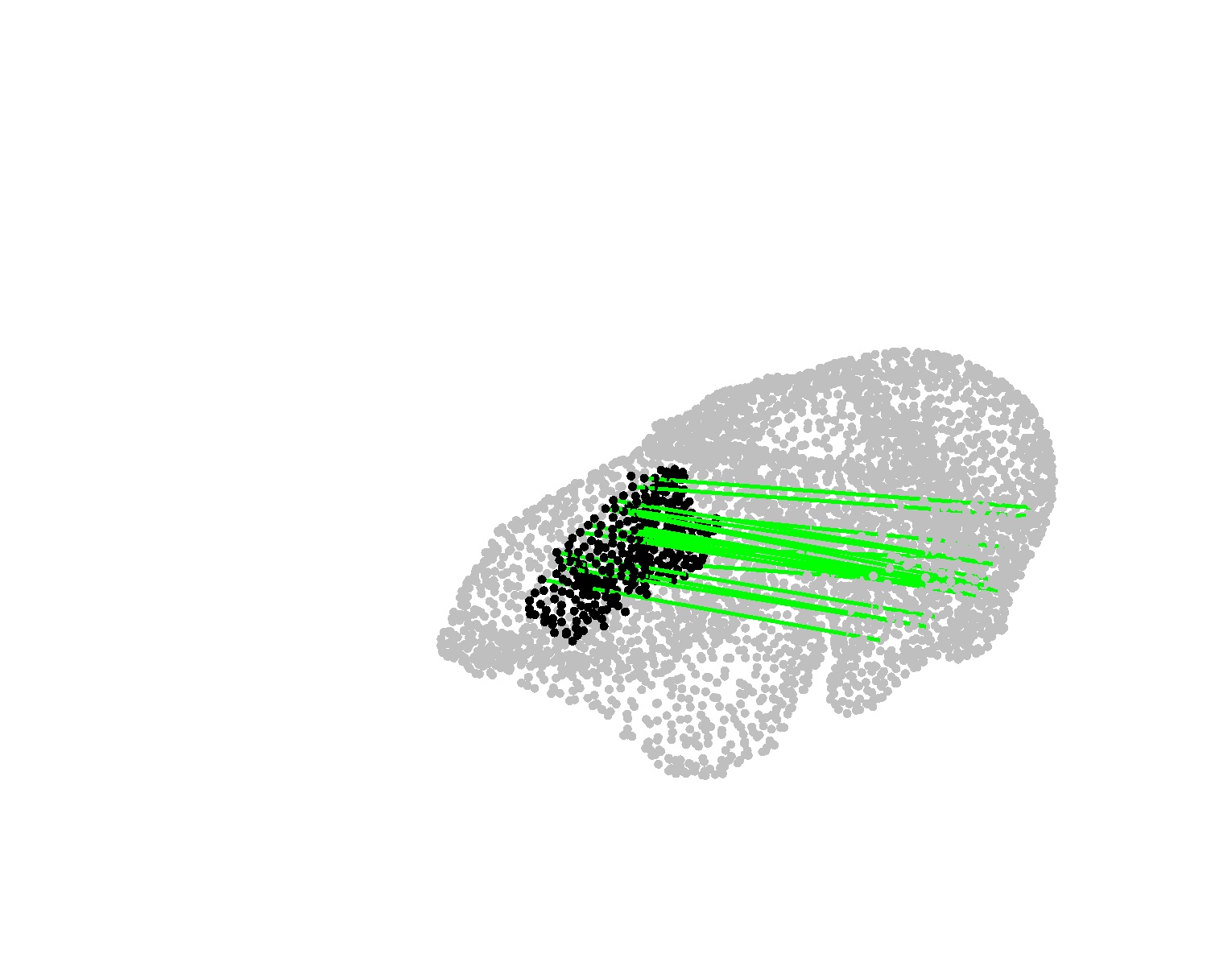}}
\hfill
\subfloat{\includegraphics[clip, trim=200 250 400 350, width=0.18\linewidth]{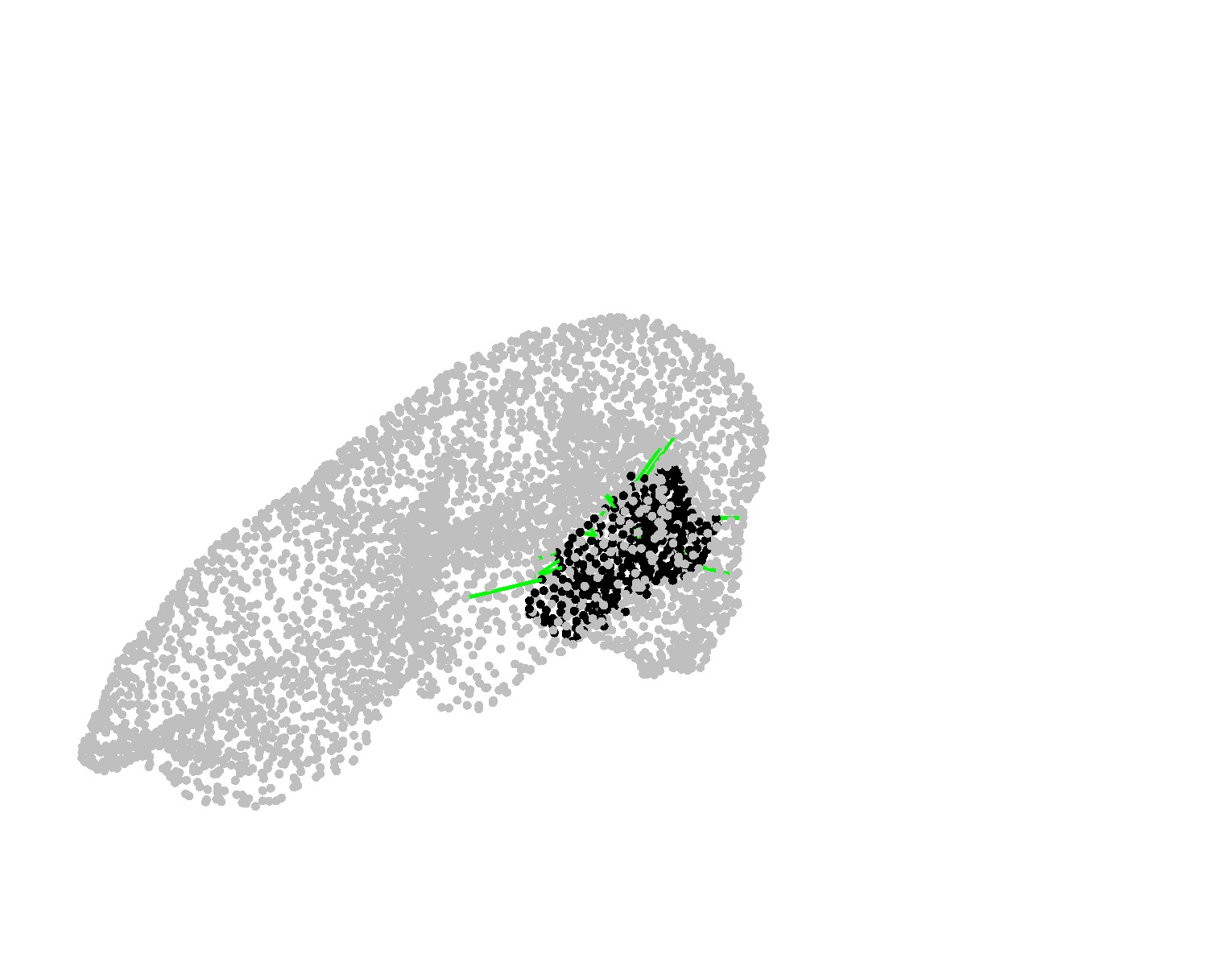}}
\\
\subfloat{\includegraphics[clip, trim=200 0 400 0, width=0.12\linewidth]{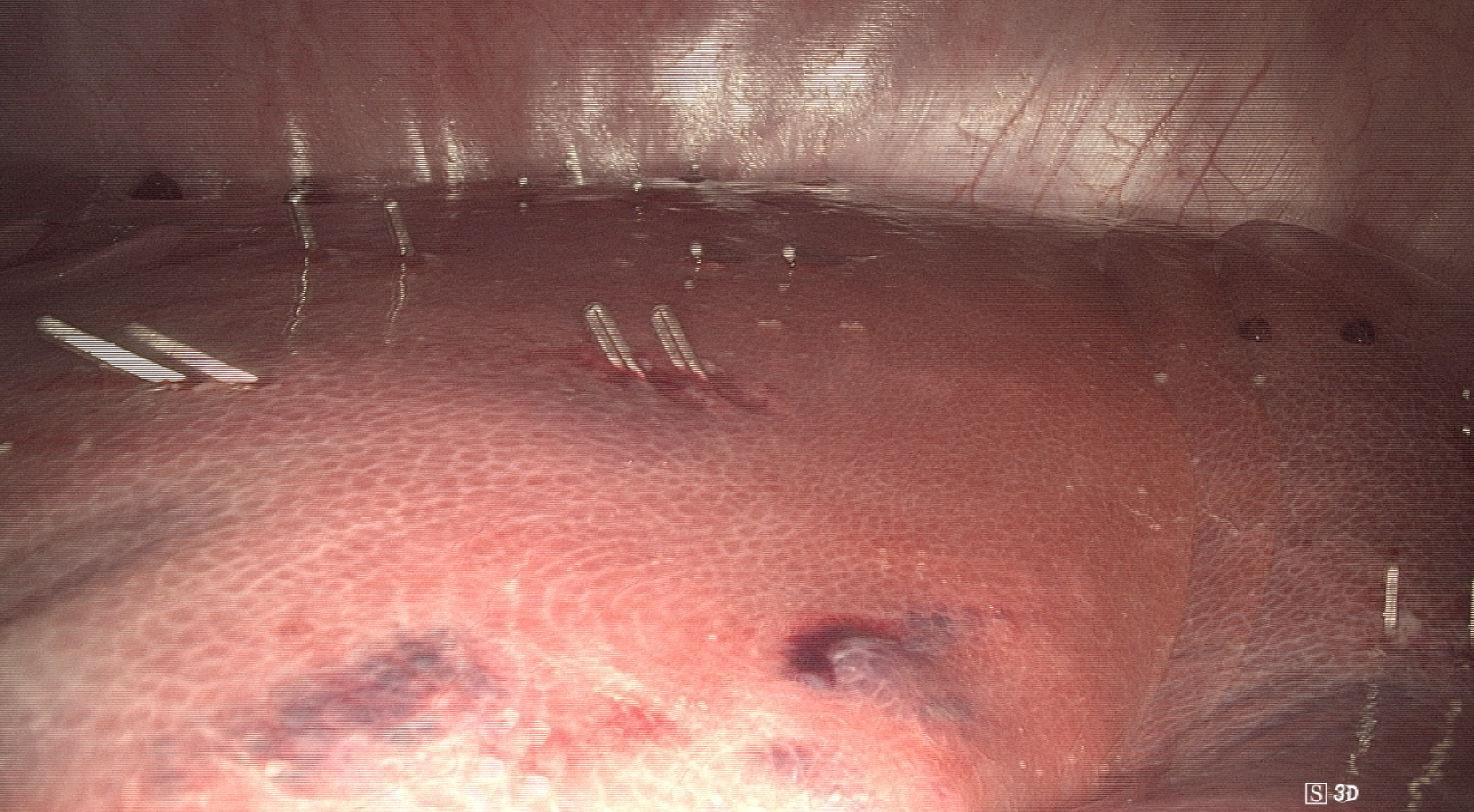}}
\hfill
\subfloat{\includegraphics[clip, trim=300 350 300 300, width=0.18\linewidth]{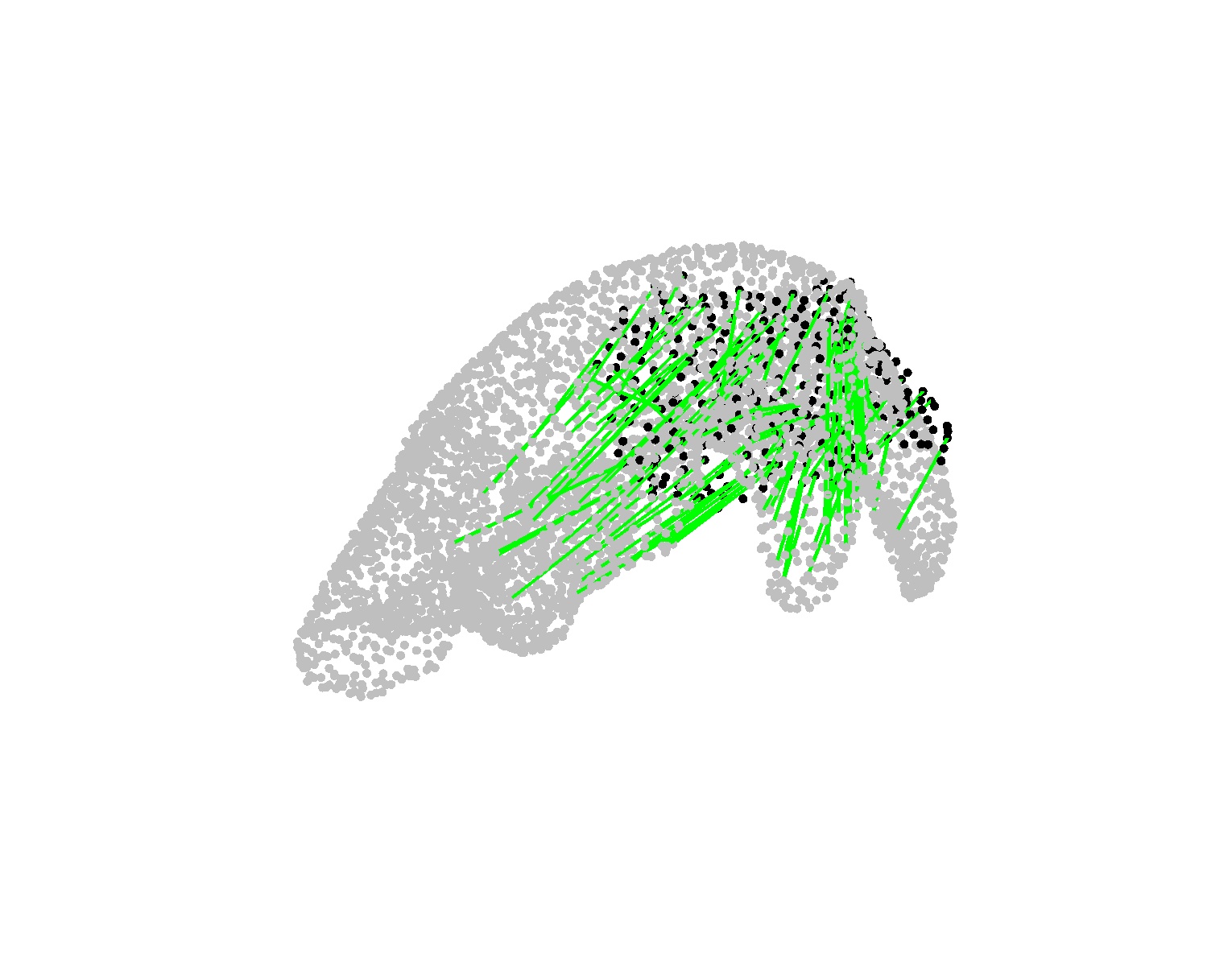}}
\hfill
\subfloat{\includegraphics[clip, trim=400 400 200 200, width=0.18\linewidth]{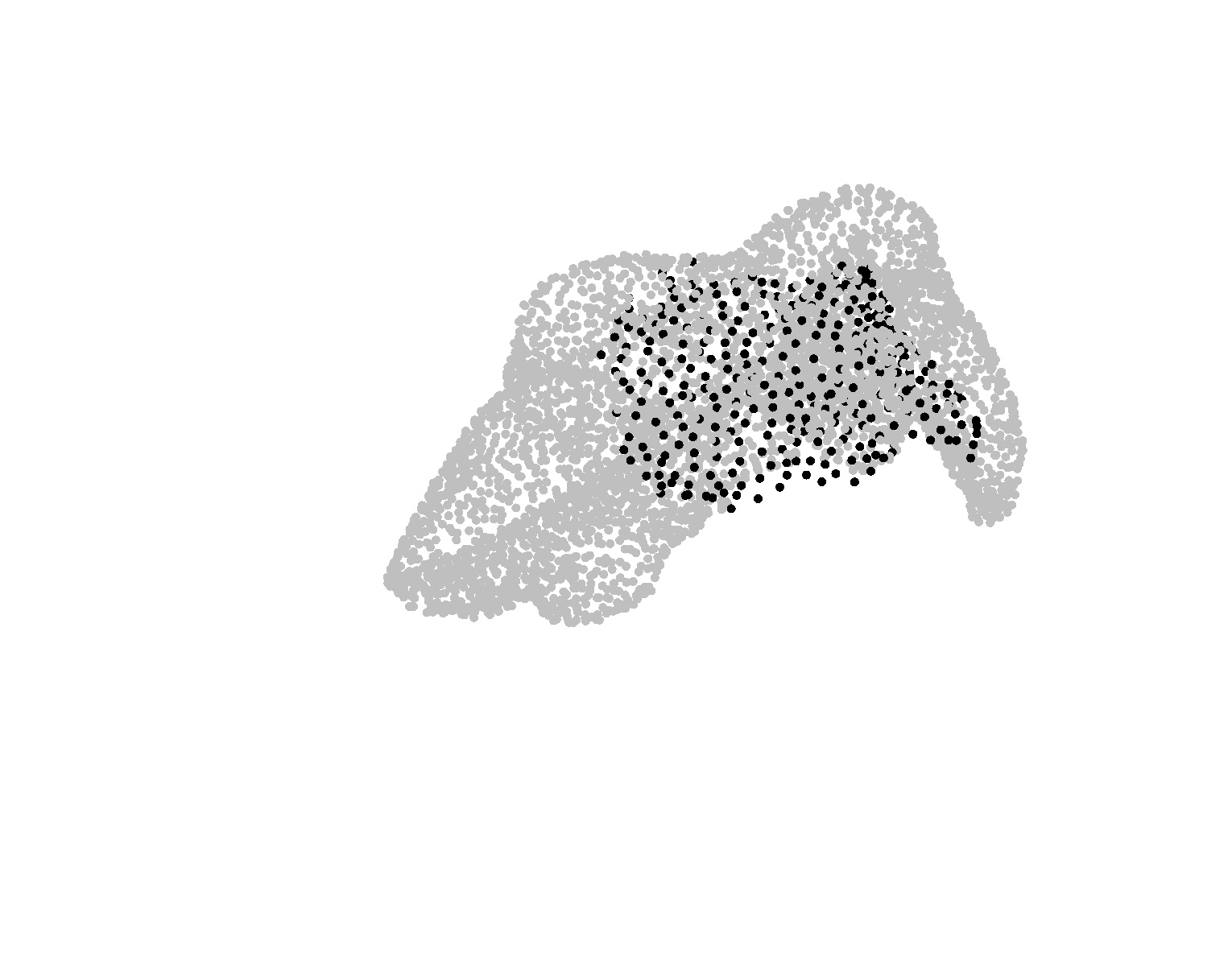}}
\hfill
\subfloat{\includegraphics[clip, trim=200 0 400 0, width=0.12\linewidth]{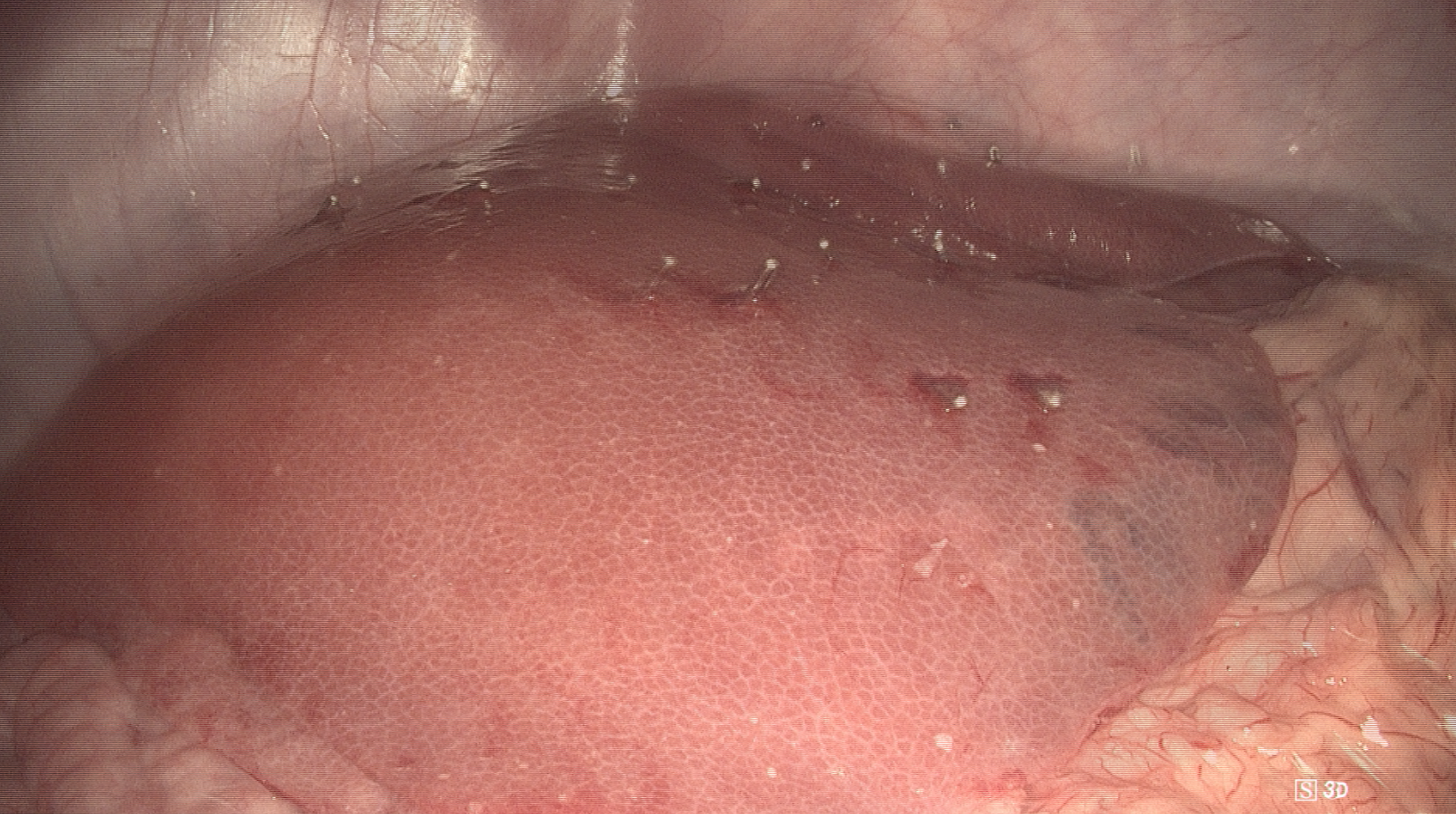}}
\hfill
\subfloat{\includegraphics[clip, trim=300 300 300 300, width=0.18\linewidth]{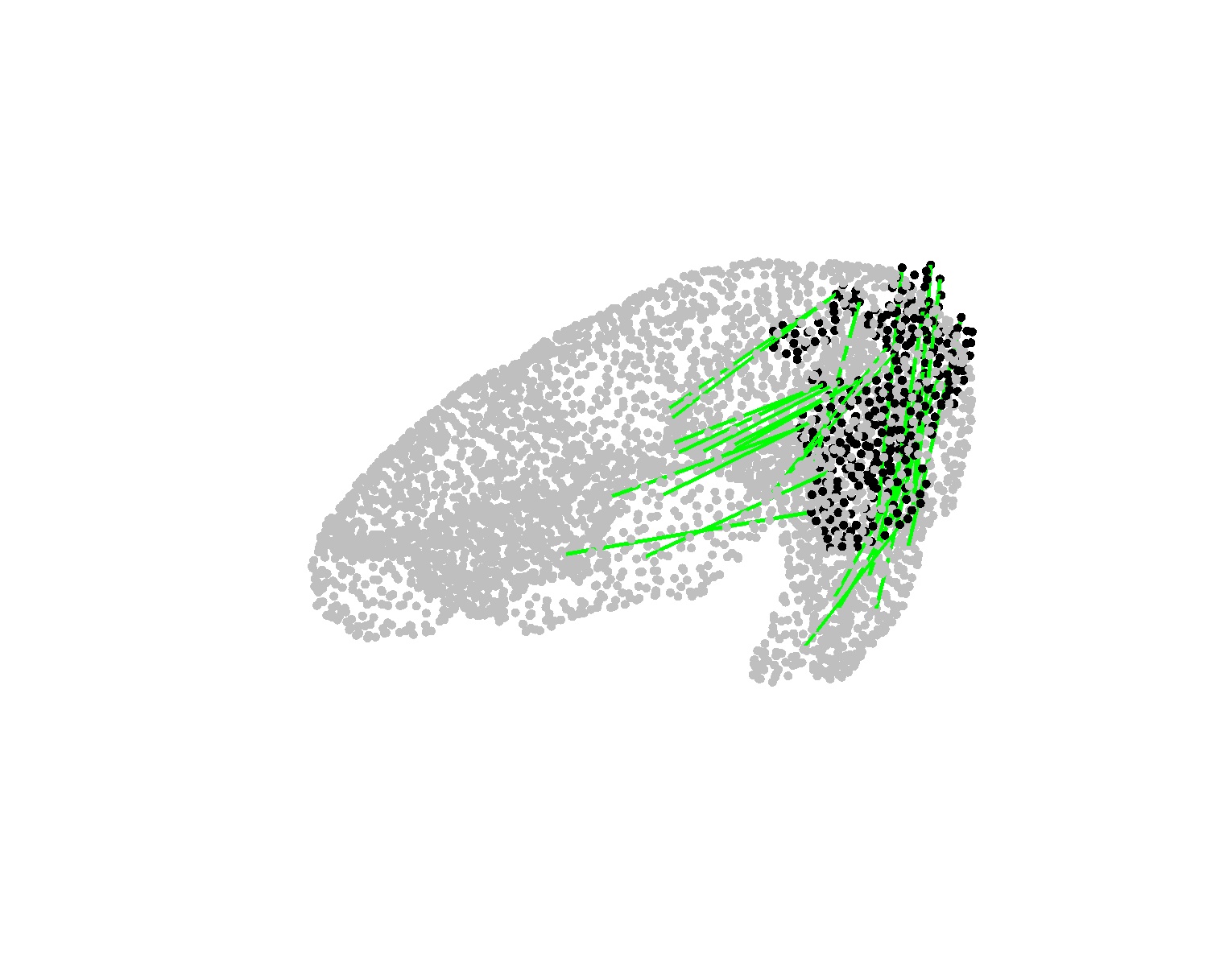}}
\hfill
\subfloat{\includegraphics[clip, trim=300 100 000 100, width=0.18\linewidth]{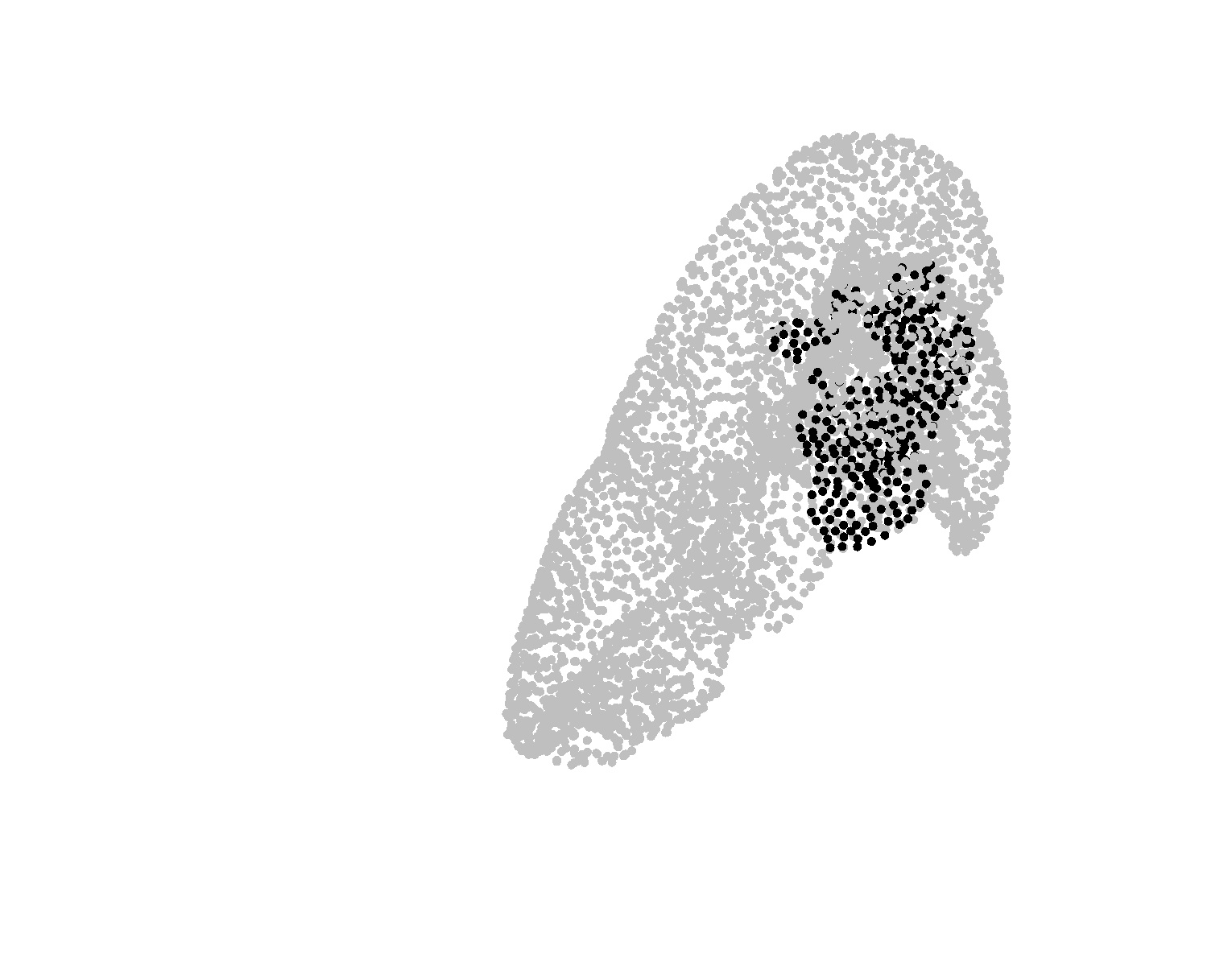}}
\caption{
Qualitative results of non-registration on six cases from the DePoLL dataset are presented, showing the endoscopic image, the matching, and the registration from left to right. Preoperative point clouds are represented in gray, intraoperative point clouds in black, and correspondences are indicated with green lines.
}
\label{fig:depoll}
\end{figure}

We then tested our method on the DePoLL porcine dataset \cite{modrzejewski2019vivo} where 3D intraoperative point clouds reconstructed from the endoscopic images. 
We use a Young's modulus of 3 KPa and a Poisson ratio of 0.45. 
We report in Table \ref{tab:registration} registration results on 13 cases from DePoll dataset.
We obtain an average Hausdorff distance (HD) of $8.45 \pm 3.60$ mm and an average FRE of $15.90 \pm 6.41$ mm, suggesting a good surface-to-surface registration and good matching.  
Visualizations of six cases are provided in Figure \ref{fig:depoll}. It is important to note that this test differs from those proposed in other works \cite{zhang2024point}, as it involves registering the reconstructed point cloud with the preoperative one, rather than with the reconstruction from the intraoperative CT scan.

\begin{table}[h!]
\centering
\caption{Registration results on DePoLL dataset (results in mm).}
\label{tab:registration}
\resizebox{1\textwidth}{!}
{
\begin{tabular}{l|cccccccccccccc}
\toprule
\textbf{Met.}  & Case0 & Case1 & Case2 & Case3 & Case4 & Case8 & Case10 & Case5 & Case6 & Case7 & Case9 & Case11 & Case12 &  Avg \\
\midrule
HD  & $11.12$ & $8.44$ & $6.62$ & $6.23$ & $15.45$ & $10.17$ & $12.04$ & $8.12$ & $5.65$ & $6.93$ & $6.88$ & $5.98$ & $7.75$ & $8.45 \pm 3.60$ \\
\rowcolor{gray!10}
FRE & $27.68$ & $18.68$ & $12.53$ & $12.23$ & $32.98$ & $26.68$ & $23.02$ & $10.43$ & $9.46$ & $15.04$ & $8.84$ & $10.54$ & $9.83$ &  $15.90 \pm 6.41$ \\
\bottomrule
\end{tabular}
}
\end{table}

\section{Conclusion}
We presented a novel method for intraoperative point cloud registration. By leveraging a patient-specific partial and deformable data generation strategy and training a Transformer-based matching network, our approach outperforms traditional agnostic methods, demonstrating improved accuracy and robustness, on synthetic and real surgical dataset.
Future work will tackle registration with topological changes and continuous registration during surgery. 

\bibliographystyle{splncs04}
\bibliography{bib}


\end{document}